CrossMark

# Shape Retrieval of Non-rigid 3D Human Models

D. Pickup[1] · X. Sun[1] · P. L. Rosin[1] · R. R. Martin[1] · Z. Cheng[2] · Z. Lian[3] ·
M. Aono[4] · A. Ben Hamza[5] · A. Bronstein[6] · M. Bronstein[7] · S. Bu[8] ·
U. Castellani[9] · S. Cheng[8] · V. Garro[9,10] · A. Giachetti[9] · A. Godil[11] · L. Isaia[9] ·
J. Han[8] · H. Johan[12] · L. Lai[13] · B. Li[14] · C. Li[15] · H. Li[13] · R. Litman[6] ·
X. Liu[13] · Z. Liu[8] · Y. Lu[14] · L. Sun[13] · G. Tam[16] · A. Tatsuma[4] · J. Ye[17]

**Abstract** 3D models of humans are commonly used within
computer graphics and vision, and so the ability to distinguish
between body shapes is an important shape retrieval problem.
We extend our recent paper which provided a benchmark for
testing non-rigid 3D shape retrieval algorithms on 3D human
models. This benchmark provided a far stricter challenge than
previous shape benchmarks. We have added 145 new models
for use as a separate training set, in order to standardise the
training data used and provide a fairer comparison. We have

✉ D. Pickup
 pickupd@cardiff.ac.uk

[1] Cardiff University, Cardiff, UK

[2] Avatar Science (Hunan) Company, Changsha, China

[3] Peking University, Beijing, China

[4] Toyohashi University of Technology, Toyohashi, Japan

[5] Concordia University, Montreal, Canada

[6] Tel-Aviv University, Tel Aviv, Israel

[7] University of Lugano, Lugano, Switzerland

[8] Northwestern Polytechnical University, Fremont, China

[9] University of Verona, Verona, Italy

[10] ISTI-CNR, Pisa, Italy

[11] National Institute of Standards and Technology, Gaithersburg, USA

[12] Fraunhofer IDM@NTU, Singapore, Singapore

[13] Beijing Technology and Business University, Beijing, China

[14] Texas State University, San Marcos, USA

[15] Duke University, Durham, USA

[16] Swansea University, Swansea, UK

[17] Penn State University, State College, USA

also included experiments with the FAUST dataset of human
scans. All participants of the previous benchmark study have
taken part in the new tests reported here, many providing
updated results using the new data. In addition, further participants
have also taken part, and we provide extra analysis
of the retrieval results. A total of 25 different shape retrieval
methods are compared.

## 1 Introduction

The ability to recognise a deformable object's shape, regardless
of the pose of the object, is an important requirement in
shape retrieval. When evaluated on previous benchmarks. the
highest performing methods achieved perfect nearest neighbour
accuracy (Lian et al. 2011, 2015), making it impossible
to demonstrate an improvement in approaches for this measure.
There is also a need for a greater variety of datasets
for testing retrieval methods, so that the research community
don't tune their methods for one particular set of data. We
recently addressed this by producing a challenging dataset
for testing non-rigid 3D shape retrieval algorithms (Pickup
et al. 2014). This dataset only contained human models, in a
variety of body shapes and poses. 3D models of humans are
commonly used within computer graphics and vision, and so
the ability to distinguish between human subjects is an important
shape retrieval problem. The shape differences between
humans are much more subtle than the differences between
the shape classes used in earlier benchmarks (e.g. various
different animals), yet humans are able to visually recognise
specific individuals. Successfully performing shape retrieval
on a dataset of human models is therefore an extremely chal-lenging, yet relevant task. Datasets of 3D humans have also been used in other tasks such as pose estimation (Ionescu et al. 2014), finding correspondences (Bogo et al. 2014), and statistical modelling (Hasler et al. 2009). For our work, the participants submitted retrieval results for a variety of methods for our human dataset, and we compared with the results in (Pickup et al. 2014). A weakness of that work is that a training set was not provided, and therefore some participants performed supervised training or parameter optimisation on the test data itself. It is therefore difficult to fairly compare the different retrieval results.

We thus provide an extension to our workshop paper (Pickup et al. 2014).[1] Firstly, participants were given 145 new human models for use as a training set. All participants who performed supervised training or parameter optimisation on the original test set retrained their method on the new training data, producing a new set of results, allowing their fairer comparison. Secondly, we have included experiments on the *FAUST* dataset (Bogo et al. 2014). Thirdly, additional participants took part in the latest tests reported here, and existing participants submitted updated or additional methods. We compare a total of 25 different retrieval methods, whereas we previously compared 21. Finally, we provide a more detailed analysis of the retrieval results.

Our paper is structured as follows. Section 2 describes the datasets used, Sect. 3 describes the retrieval task, Sect. 4 outlines all methods tested, organised by submitting participant, Sect. 5 provides a detailed analysis of the retrieval results, and finally we conclude in Sect. 6.

## 2 Datasets

The human models we use are split into three datasets. The first two datasets, which we created ourselves, consist of a *Real* dataset, obtained by scanning real human participants and generating synthetic poses, and a *Synthetic* dataset, created using 3D modelling software (DAZ 2013). The latter may be useful for testing algorithms intended to retrieve synthetic data, with well sculpted local details, while the former may be more useful to test algorithms that are designed to work even in the presence of noisy, coarsely captured data lacking local detail. The third dataset we use is the *FAUST* dataset created by Bogo et al. (2014), which uses scans of different people, each in a set of different poses, and contains both topological noise and missing parts.

Our *Real* and *Synthetic* datasets are available to download from our benchmark website (Footnote 1), or from the

doi:10.17035/d.2015.100097. The *FAUST* dataset is available from its project website.[2]

Throughout the paper we use the following terms when referring to our data:

Model —A single 3D object.
Mesh —The underlying triangle mesh representation of a model.
Subject —A single person. The datasets' models are divided into classes, one class for each subject.
Pose —The articulation or conformation of a model (e.g. standing upright with arms by the sides).
Shape —The pose-invariant form of a model (i.e. aspects of the model shape invariant to pose).

### 2.1 Real Dataset

The *Real* dataset was built from point-clouds contained within the Civilian American and European Surface Anthropometry Resource (CAESAR) (CAESAR 2013). The original *Test* set contained 400 models, representing 40 human subjects (20 male, 20 female), each in ten different poses. The poses we used are a random subset of the poses used for the SCAPE (Anguelov et al. 2005) dataset. The same poses were used for each subject. Our new *Training* set contains 100 models, representing 10 human subjects (5 male, 5 female), again in 10 different poses. None of the training subjects or poses are present in the test set (Fig. 1).

The point-clouds were manually selected from CAESAR to have significant visual differences. We employed SCAPE (shape completion and animation of people) (Anguelov et al. 2005) to build articulated 3D meshes, by fitting a template mesh to each subject (Fig. 2). Realistic deformed poses of each subject were built using a data-driven deformation technique (Chen et al. 2013). We remeshed the models using freely available software (Valette and Chassery 2004; Valette et al. 2008) so different meshes do not have identical triangulations. As the same remeshing algorithm was applied to all meshes, the triangulations may share similar properties, but exact correspondences cannot be derived directly from the vertex indices of the meshes. The resulting meshes each have approximately 15,000 vertices, varying slightly from mesh to mesh.

While we used a data-driven technique to generate the poses, generating them synthetically means they do not exhibit as realistic surface deformations between poses as different scans would have done. The data also does not suffer from missing parts or topological noise sometimes found in scanned data. A selection of models from this dataset is shown in Fig. 1a.

---

[1] Benchmark Website: http://www.cs.cf.ac.uk/shaperetrieval/shrec14/.

[2] FAUST Website: http://faust.is.tue.mpg.de/.

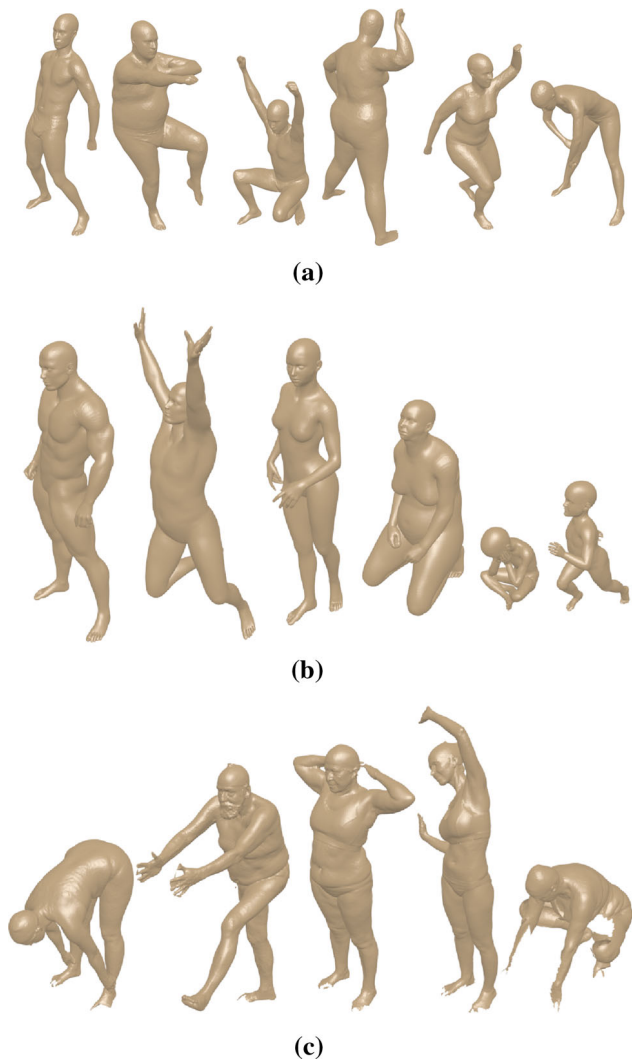

**Fig. 1** A selection of models included in the datasets. **a** Real dataset. **b** Synthetic dataset. **c** FAUST dataset

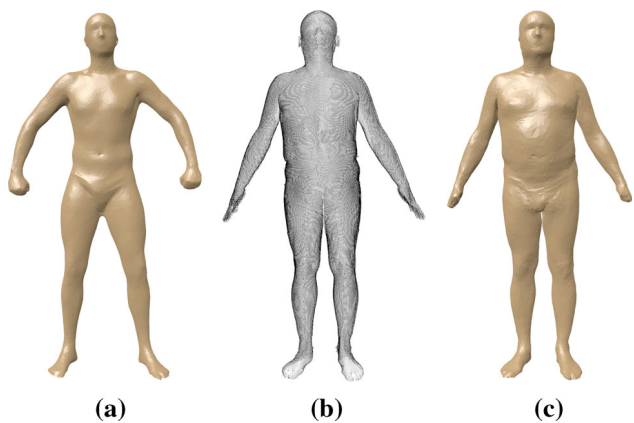

**Fig. 2** A template mesh is fitted to each point cloud scan using the SCAPE method (Anguelov et al. 2005). **a** Template mesh. **b** Point cloud. **c** Template fitted to point cloud

## 2.2 Synthetic Dataset

We used the DAZ Studio (DAZ 2013) 3D modelling and animation software to create a dataset of synthetic human models. The software includes a parameterized human model, where parameters control body shape. We used this to produce a *Test* dataset consisting of 15 different human subjects (5 male, 5 female, 5 child), each with its own unique body shape. We generated 20 different poses for each model, resulting in a dataset of 300 models. The poses were chosen by hand from a palette of poses provided by DAZ Studio. The poses available in this palette contain some which are simple variations of each other, so we therefore hand picked poses representing a wide range of articulations. The same poses were used for each subject. Our new *Training* set contains 45 models, representing 9 human subjects (3 male, 3 female, 3 child) in 5 different poses. None of the training subjects or poses is present in the test set. All models were remeshed, as for the *Real* dataset. The resulting meshes have approximately 60,000 vertices, again varying slightly. A selection of these models is shown in Fig. 1b.

## 2.3 FAUST Dataset

The *FAUST* dataset was created by scanning human subjects with a sophisticated 3D stereo capture system. The *Test* dataset consists of 10 different human subjects, with each subject being captured in the same 20 poses, resulting in a dataset of 200 models. The *Training* set contains 100 models, made up of 10 subjects in 10 poses. The average number of vertices is 172,000, making it the highest resolution of the three datasets. A selection of models from this dataset is shown in Fig. 1c.

As the poses for this dataset were generated from scans, they contain realistic deformations that are normally missing from synthetic models. The models also have missing parts caused by occlusion, and topological noise where touching body parts are fused together. The dataset also contains some non-manifold vertices and edges, which some retrieval methods cannot handle. We therefore produced a version of the data from which these non-manifold components were removed and holes filled, creating a watertight manifold for each model. This mesh processing was performed using Meshlab (MeshLab 2014), and the same automatic process was applied to all meshes. There was no hand-correction of any of the results of this procedure. Apart from these small local changes, the data was otherwise unmodified. Some examples of the watertight meshes are shown in Fig. 3. Our watertight models were distributed to participants upon request. For the full details of the *FAUST* dataset we refer readers to Bogo et al. (2014).

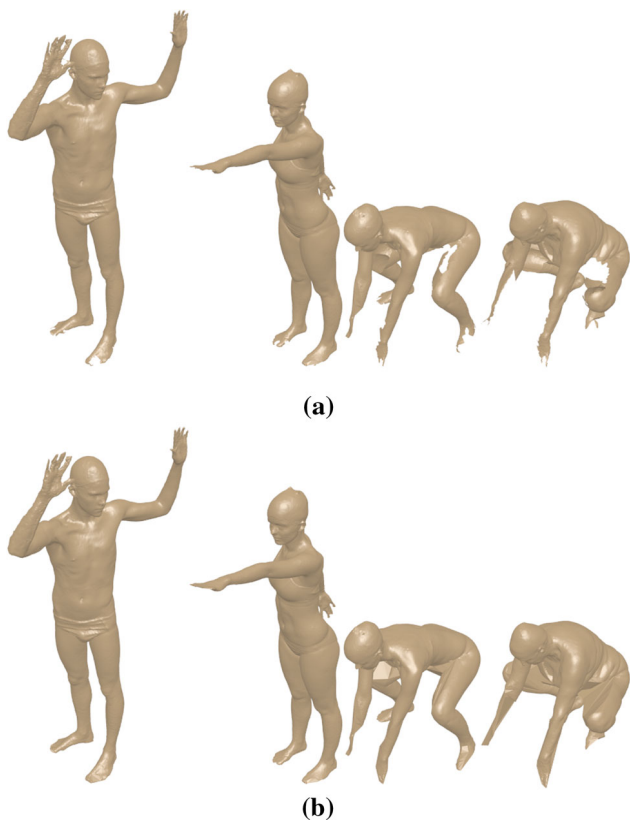

**(a)**

**(b)**

**Fig. 3** Examples of the watertight *FAUST* meshes. **a** Original meshes with missing data. **b** Watertight manifold versions produced by Meshlab

**Table 1** Summary of methods, including details of any mesh simplification and use of watertight meshes for the FAUST dataset

| Author | Method | Simplification | Watertight (FAUST) |
|---|---|---|---|
| Giachetti | APT | No | Used |
| | APT-trained | No | Used |
| Lai | HKS | 10,000 faces | Used |
| | WKS | 10,000 faces | Used |
| | SA | 10,000 faces | Used |
| | Multi-feature | 10,000 faces | Used |
| B. Li | Curvature | No | Used |
| | Geodesic | 1000 vertices | Used |
| | Hybrid | 1000 vertices | Used |
| | MDS-R | 1000 vertices | Used |
| | MDS-ZFDR | 1000 vertices | Used |
| C. Li | Spectral Geom. | No | Used |
| Litman | supDL | 4500 vertices | Used |
| | UnSup32 | 4500 vertices | Used |
| | softVQ48 | 4500 vertices | Used |
| Pickup | Surface area | No | Used |
| | Compactness | No | Used |
| | Canonical | No | Used |
| Bu | 3DDL | No | Used |
| Tatsuma | BoF-APFH | No | Not used |
| | MR-BoF-APFH | No | Not used |
| Ye | R-BiHDM | No | Used |
| | R-BiHDM-s | No | Used |
| Tam | MRG | No | Used |
| | TPR | No | Used |

## 3 Retrieval Task and Evaluation

All participants in our study submitted results for the following retrieval task:

> Given a query model, return a list of all models, ordered by decreasing shape similarity to the query.

Every model in the database was used in turn as a separate query model.

The evaluation procedure used to assess the results (see Sect. 5) is similar to that used by previous comparative studies (Lian et al. 2011, 2015). We evaluate the results using various statistical measures: nearest neighbour (NN), first tier (1-T), second tier (2-T), e-measure (E-M), discounted cumulative gain (DCG), and precision and recall curves. Definitions of these measures are given in Shilane et al. (2004).

## 4 Methods

We now briefly describe each of the methods compared in our study; as can be seen, some participants submitted multiple methods. Table 1 summarised which methods simplified the meshes to a lower resolution, and which used the watertight version of the *FAUST* dataset. Approximate timings of each method are given in Table 2. Full details of these methods may be found in the papers cited.

### 4.1 Simple Shape Measures, and Skeleton Driven Canonical Forms

This section presents two techniques, simple shape measures based on simple invariant intrinsic geometric properties, and skeleton driven canonical forms.

#### 4.1.1 Simple Shape Measures

We may observe that to a good approximation, neither the surface area nor the volume of the model should change under deformation. The first measure is thus the total surface area $A$ of the mesh. This measure is not scale independent, and all human models were assumed to be properly scaled. In

**Table 2** Approximate timing information for preprocessing and computing the model descriptors per model for each method

| Author | Method | Timings | | | | | | Language |
|---|---|---|---|---|---|---|---|---|
| | | Real | | Synthetic | | FAUST | | |
| | | Preprocessing (s) | Model descriptor (s) | Preprocessing (s) | Model descriptor (s) | Preprocessing (s) | Model descriptor (s) | |
| Giachetti | APT | – | 38 | – | 25 | – | 40 | C++ |
| | APT-trained | 3810 | 38 | 1135 | 25 | 4010 | 40 | C++ |
| Lai | HKS | – | 15 | – | 17 | – | 15 | Matlab |
| | WKS | – | 12 | – | 13 | – | 12 | Matlab |
| | SA | – | <1 | – | <1 | – | <1 | Matlab |
| | Multi-feature | 2708 | 27 | 1356 | 29 | 2709 | 27 | Matlab |
| B. Li | Curvature | – | 14 | – | 57 | – | 4 | C++ |
| | Geodesic | – | 51 | – | 54 | – | – | Matlab, C++ |
| | Hybrid | 48,200 | 119 | 54,884 | 178 | – | – | Matlab, C++ |
| | MDS-R | – | 54 | – | 67 | – | – | Matlab, C++ |
| | MDS-ZFDR | – | 54 | – | 67 | – | – | Matlab, C++ |
| C. Li | Spectral Geom. | 2700 | 8 | 2700 | 37 | 2700 | 96 | Matlab |
| Litman | supDL | 18,000 | 3 | 18,000 | 3 | 18,000 | 3 | Matlab, C++ |
| | UnSup32 | 900 | 3 | 900 | 3 | 900 | 3 | Matlab, C++ |
| | softVQ48 | 3600 | 3 | 3600 | 3 | 3600 | 3 | Matlab, C++ |
| Pickup | Surface area | – | <1 | – | <1 | – | <1 | Matlab |
| | Compactness | – | <1 | – | 2 | – | 3 | Matlab |
| | Canonical | – | 20 | – | 106 | – | 510 | Matlab, C++ |
| Bu | 3DDL | 3600 | 10 | 3600 | 10 | 3600 | 10 | Matlab, C++ |
| Tatsuma | BoF-APFH | 74 | 3 | 82 | 3 | 78 | 4 | Python, C++ |
| | MR-BoF-APFH | 74 | 3 | 82 | 3 | 78 | 4 | Python, C++ |
| Ye | R-BiHDM | – | 10 | – | 40 | – | 55 | C++ |
| | R-BiHDM-s | – | 30 | – | 120 | – | 160 | C++ |
| Tam | MRG | – | 51 | – | 834 | – | – | C++ |
| | TPR | – | 11 | – | 225 | – | 1126 | C++ |

Preprocessing may refer to training, dictionary learning or high-level feature learning, please see the method descriptions for details. Methods which don't perform preprocessing have a '–' given as their preprocessing time. When supervised training is used, the time to compute any features from the training data is included in the preprocessing time given. Please note that the different methods may have been implemented in different languages and were tested on different hardware, therefore any small differences in timings are not directly comparable. *s* seconds order to account for a possibly unknown scale, the second measure, compactness $C$ uses the volume $V$ to provide a dimensionless quantity: $C = V^2/A^3$. Both measures are trivial to implement, and are very efficient to compute.

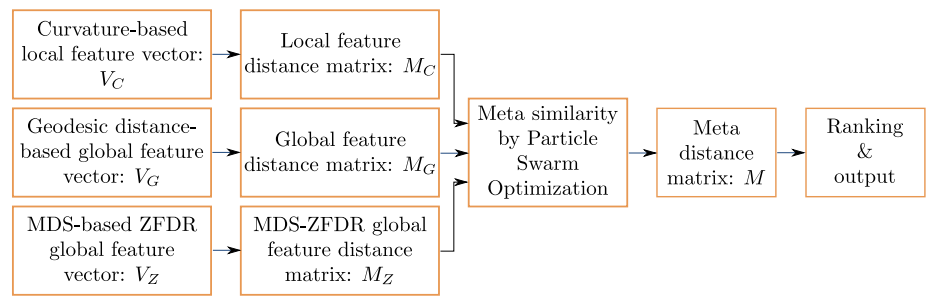

The surface area $A$ is the sum of the triangle areas:

$$A = \sum_{i=1}^{N} A_i = \frac{1}{2} \sum_{i=1}^{N} |(b_i - c_i) \times (a_i - b_i)|, \qquad (1)$$

where the $i$th triangle has vertices $(a_i, b_i, c_i)$ in anticlockwise order, $\times$ denotes vector cross-product, and $N$ is the number of triangles. The volume $V$ of the mesh is calculated as:

$$V = \frac{1}{6} \sum_{i=1}^{N} a_i \cdot (b_i \times c_i). \qquad (2)$$

We do not take into account any self-intersections occurring in the meshes, and therefore the volume calculation may not be accurate for certain certain poses; this is a weakness of this simple method.

### 4.1.2 Skeleton Driven Canonical Forms

This method uses a variant of the canonical forms presented by Elad and Kimmel (2003) to normalise the pose of all models in the dataset, and then uses the rigid view-based method in Lian et al. (2013a) for retrieval. This method works as follows (Pickup et al. 2016). A canonical form is produced by extracting a curve skeleton from a mesh, using the method in Au et al. (2008). The SMACOF multidimensional scaling method used in Elad and Kimmel (2003) is then applied to the skeleton, to put the skeleton into a canonical pose. The skeleton driven shape deformation method in Yan et al. (2008) is then used to deform the mesh to the new pose defined by the canonical skeleton. This produces a similar canonical form to the one in Elad and Kimmel (2003), but with local features better preserved, similarly to Lian et al. (2013b).

The retrieval method by Lian et al. (2013a) performs retrieval using the canonical forms by rendering a set of 66 depth views of each object, and describing each view using *bag-of-features*, with *SIFT* features. Each pair of models is compared using the *bag-of-features* descriptors of their associated views.

In Pickup et al. (2014) the *Synthetic* models had to be simplified, but we have now made some minor coding improvements which allows the method to run on the full resolution meshes for all three datasets.

### 4.2 Hybrid Shape Descriptor and Meta Similarity Generation for Non-rigid 3D Model Retrieval

The hybrid shape descriptor in (Li et al. 2014) integrates both geodesic distance-based global features and curvature-based local features. An adaptive algorithm based on *particle swarm optimization* (PSO) is developed to adaptively fuse different features to generate a meta similarity between any two models. The approach can be generalized to similar approaches which integrate more or different features. Figure 4 shows the framework of the hybrid approach. It first extracts three component features of the hybrid shape descriptor: curvature-based local features, geodesic distance-based global features, and multidimensional scaling (MDS) based ZFDR global features (Li and Johan 2013). Based on these features, corresponding distance matrices are computed and fused into a meta-distance matrix based on PSO. Finally, the distances are sorted to generate the retrieval lists.

#### 4.2.1 Curvature-based local feature vector: $V_C$

First, a *curvature index* feature is computed to characterise local geometry for each vertex $p$:

$$CI = \frac{2}{\pi} \log(\sqrt{(K_1^2 + K_2^2)/2}),$$

where $K_1$ and $K_2$ are two principal curvatures at $p$. Then, a curvature index deviation feature is computed for vertices adjacent to $p$:

$$\delta CI = \sqrt{(\sum_{i=1}^{n}(CI_i - \widetilde{CI})^2)/n},$$

where $CI_1, \ldots, CI_n$ are the curvature index values of adjacent vertices and $\widetilde{CI}$ is the mean curvature index for all adjacent vertices. Next, the *shape index* feature for describing local topology at $p$ is computed as

$$SI = \frac{2}{\pi} \arctan((K_1 + K_2)/|K_1 - K_2|).$$

A combined local shape descriptor is then formed by concatenating these local features: $F = (CI, \delta CI, SI)$. Finally, based on the bag-of-words framework, the local feature vector $V_C = (h_1, \ldots, h_{N_C})$ is formed, where the number of cluster centres $N_C$ is set to 50.

### 4.2.2 Geodesic Distance-Based Global Feature Vector: $V_G$

To avoid the high computational cost of computing geodesic distances between all vertices, each mesh is first simplified to 1000 vertices. The geodesic distance between each pair of its vertices is then computed to form a geodesic distance matrix, which is then decomposed using singular value decomposition. The ordered largest $k$ singular values form a global feature vector. Here, $k = 50$.

### 4.2.3 MDS-Based ZFDR Global Feature Vector: $V_Z$

To create a pose invariant representation of non-rigid models, MDS is used to map the non-rigid models into a 3D canonical form. The geodesic distances between the vertices of each simplified 3D model are used as the input to MDS for feature space transformation. Finally, the hybrid global shape descriptor ZFDR (Li and Johan 2013) is used to characterize the features of the transformed 3D model in the new feature space. There are four feature components in ZFDR: Zernike moments, Fourier descriptors, Depth information and Ray-based features. This approach is called MDS-ZFDR, stressing that MDS is adopted in the experiments. For 3D human retrieval, using the R feature only (that is MDS-R) always achieves better results than other combinations such as ZF, DR or ZFDR. This is because salient feature variations in the human models, e.g. fat versus slim, are better characterised by the R feature than other visual-related features like Z, F and D.

### 4.2.4 Retrieval Algorithm

The complete retrieval process is as follows:

1. Compute curvature-based local feature vector $V_C$ based on the original models and generate local feature distance matrix $M_C$.
2. Compute geodesic distance-based global feature vector $V_G$ and global feature distance matrix $M_G$.

3. Compute MDS-based ZFDR global feature vector $V_Z$ and MDS-ZFDR global feature distance matrix $M_Z$.
4. Perform PSO-based meta-distance matrix generation as follows:

The meta-distance matrix $M = w_C M_C + w_G M_G + w_Z M_Z$ depends on weights $w_C$, $w_G$ and $w_Z$ in [0,1]. The weights used in this paper were obtained by training the above retrieval algorithm using the PSO algorithm on the training dataset: for the *Real* dataset, $w_C = 0.7827$, $w_G = 0.2091$ and $w_Z = 0.0082$; for the *Synthetic* dataset, $w_C = 0.4416$, $w_G = 0.5173$ and $w_Z = 0.0410$.

As a swarm intelligence optimization technique, the PSO-based approach can robustly and quickly solve nonlinear, non-differentiable problems. It includes four steps: initialization, particle velocity and position updates, search evaluation and result verification. The number of particles used is $N_P = 10$, and the maximum number of search iterations is $N_t = 10$. The first tier is selected as the fitness value for search evaluation. Note that the PSO-based weight assignment preprocessing step is only performed once on each training dataset.

### 4.3 Histograms of Area Projection Transform

This approach uses *histograms of area projection transforms* (HAPT), general purpose shape descriptors proposed in Giachetti and Lovato (2012), for shape retrieval. The method is based on a spatial map (the *multiscale area projection transform*) that encodes the likelihoods that 3D points inside the mesh are centres of spherical symmetry. This map is obtained by computing for each radius of interest the value:

$$\text{APT}(\mathbf{x}, S, R, \sigma) = \text{Area}(T_R^{-1}(k_\sigma(\mathbf{x}) \subset T_R(S, \mathbf{n}))), \quad (3)$$

where $S$ is the surface of interest, $T_R(S, \mathbf{n})$ is the parallel surface to $S$ shifted (inwards only) along the normal vector $\mathbf{n}$ by a distance $R$, $T_R^{-1}$ is the part of the original surface used to generate the parallel surface $T_R$, and $k_\sigma(\mathbf{x})$ is a sphere of radius $\sigma$ centred on the generic 3D point $\mathbf{x}$ where the map is computed (Fig. 5). Values at different radii are normalized to provide scale-invariant behaviour, creating the multiscale APT (MAPT):

$$\text{MAPT}(\mathbf{x}, R, S) = \alpha(R)\, \text{APT}(\mathbf{x}, S, R, \sigma(R)), \quad (4)$$

where $\alpha(R) = 1/4\pi R^2$ and $\sigma(R) = cR$, $(0 < c < 1)$.

The discretized MAPT is easily computed, for selected values of $R$, on a voxelized grid containing the surface mesh by the procedure in Giachetti and Lovato (2012). The map is computed on a grid of voxels of size $s$ on a set of corresponding sampled radius values $R_1, \ldots, R_n$. Histograms

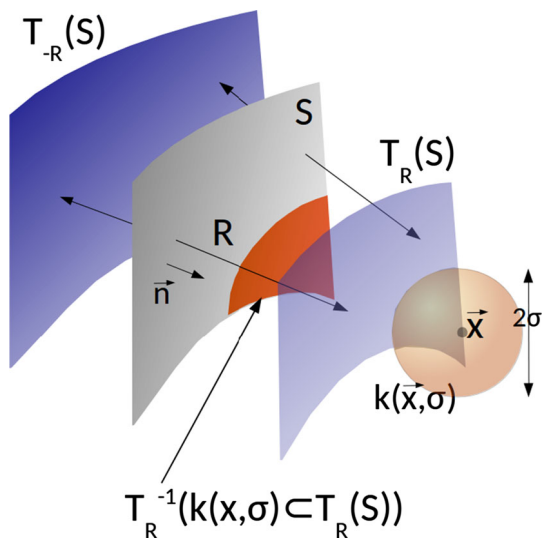

**Fig. 5** Basic idea of the area projection transform: we compute the parallel surface at distance R and we compute the transform at a point **x** as the area of the original surface generating the part of the parallel surface falling inside a sphere of radius $\sigma$ centred at **x**

of MAPT computed inside objects are good global shape descriptors, as shown by their very good performance on the SHREC'11 non-rigid watertight contest data (Lian et al. 2011). For that recognition task, discrete MAPT maps were quantized in 12 bins and histograms computed at the selected radii were concatenated to create a descriptor. Voxel side and sampled radii were chosen, proportional to the cube root of the object volume for each model, to normalize the descriptor independently of scale. The parameter $c$ was set to 0.5.

To recognise human subjects, however, scale invariance is not desired. For this reason a fixed voxel size and a fixed set of radii is used. The values for these parameters were chosen differently for each dataset, by applying simple heuristics to the training data. For all datasets, the MAPT maps were quantized into 6 bins. The voxel size was taken to be similar to the size of the smaller well defined details in the meshes. For the *Synthetic* dataset, where fingers are clearly visible and models are smaller, $s = 4$ mm is used; the MAPT histograms are computed for 11 increasing radii starting from $R_1 = 8$ mm, in increments of 4 mm for the remaining values. In the *Real* dataset, models are bigger and details are more smoothed, so we set $s = 12$ mm and use 15 different radii starting from $R_1 = 24$ mm radius in increments of 12 mm. For the *FAUST* dataset we use the same parameters as for the *Real* dataset.

Measuring distances between models simply involves concatenating the MAPT histograms computed at different scales and evaluating the Jeffrey divergence of the corresponding concatenated vectors.

### 4.3.1 Trained Approach

The available training dataset was exploited to project the original feature space into a subspace that is maximally discriminative for different instances of the specific class of objects; distances are computed on the mapped descriptors. The mapping uses a combination of principal component analysis (PCA) and linear discriminant analysis (LDA) (Duda et al. 2012).

PCA transforms the data set into a different coordinate system in which the first coordinate in the transformed domain, called the principal component, has maximum variance and other coordinates have successively smaller variances. LDA puts a labelled dataset into a subspace which maximizes between-class scatter. The combination of these two mappings first decorrelates the data and then maximizes the variances between classes. The combined mapping is defined as: $D_{map} = LDA(PCA(D))$. Several tests indicated 10 dimensions should be used for the PCA. The dimensionality of the original descriptors is 180. Regularized LDA can be used to bypass the initial PCA computation, but we find that using PCA followed by standard LDA performs better in practice. For the mappings, the Matlab implementation in the PRTools 5 package (Van Der Heijden et al. 2005) was used. The PCA and LDA procedures are very efficient, only accounting for 10 s of the full training time given in Table 2. The rest of the time is spent computing the descriptors from the training data to be input into the PCA and LDA algorithms.

The improvements that can be obtained with this approach clearly depend on the number of examples available in the training set and how well these examples represent the differences found in the test set. The improvements are less evident for the *Synthetic* dataset, where the number of training examples is lower and we find that they do not fully characterise range of body shapes present in the test set.

### 4.4 R-BiHDM

The R-BiHDM (Ye et al. 2013; Ye and Yu 2015) method is a spectral method for general non-rigid shape retrieval. Using modal analysis, the method projects the *biharmonic distance* map (Lipman et al. 2010) into a low-frequency representation which operates on the modal space spanned by the lowest eigenfunctions of the shape Laplacian (Reuter et al. 2006; Ovsjanikov et al. 2012), and then computes its spectrum as an isometric shape descriptor.

Let $\psi_0, \ldots, \psi_m$ be the eigenfunctions of the Laplacian $\Delta$, corresponding to its smallest eigenvalues $0 = \lambda_0 \leq \ldots \leq \lambda_m$. Let $d(x, y)$ be the biharmonic distance between two points on a mesh, defined as

$$d(x, y)^2 = \sum_{i=1}^{m} \frac{1}{\lambda_i^2} \left(\psi_i(x) - \psi_i(y)\right)^2. \tag{5}$$

The squared biharmonic distance map $\mathcal{D}^2$ is a functional map defined by

$$\mathcal{D}^2[f](x) = \int_{x \in S} d^2(x, y) f(y) \mathrm{d}y, \tag{6}$$

where $S$ is a smooth manifold. The reduced matrix version of $\mathcal{D}^2$ is denoted by $A = \{a_{i,j}\}$, where $a_{i,j} = \int_S \psi_i(x) \mathcal{D}^2[\psi_j](x) \mathrm{d}x$ for $0 \le i, j \le m$. Note that $\mathrm{tr}(A) = 0$ and all eigenvalues of $A$, denoted by $\mu_0, \ldots, \mu_m$ are in descending order of magnitude, where $\mu_0 > 0$ and $\mu_i < 0$ for $i > 0$. The shape descriptor is defined by the vector $[\mu_1, \ldots, \mu_m]^T$ (for a scale dependent version) or $[\mu_1/\mu_0, \ldots, \mu_L/\mu_0]^T$ (scale independent). In this test, $L = 30$ and $m = 60$ for the scale independent version, and $L = m = 100$ for the scale dependent version. Finally, a normalized Euclidean distance is used for nearest neighbour queries. The descriptor is insensitive to a number of perturbations, such as isometry, noise, and remeshing. It has good discrimination capability with respect to global changes of shape and is very efficient to compute. We have found that the scale independent descriptor (R-BiHDM) is more reliable for generic nonrigid shape tasks, while the scale dependent descriptor (R-BiHDM-s) is more suitable for this human shape task (see Sect. 5).

### 4.5 Multi-feature Descriptor

Single feature descriptors cannot capture all aspects of a shape, so this approach fuses several features into a multifeature descriptor to improve retrieval accuracy. Three state-of-the-art features are used: *heat kernel signatures* (HKS) (Sun et al. 2009), *wave kernel signatures* (WKS) (Aubry et al. 2011) and mesh surface area (SA).

Firstly, the similarity of all the models in the training set is calculated for each of the three chosen features. Secondly, some models are selected at random to produce a subset of the training data, with the rest left for validation. For each feature $f_i$, its entropy is calculated as

$$E(f_i) = -\sum_{j=1}^{N} p_j^i \log_2 p_j^i, \tag{7}$$

where $N$ is the number of shape classes and $p_j^i$ is the probability distribution of shape class $j$ for feature $i$. A weighting for each feature is then calculated as

$$w_i = \frac{1 - E(f_i)}{3 - \sum E(f_i)}. \tag{8}$$

Having determined the weights, the combined similarity matrix $\mathbf{S}$ is calculated as

$$\mathbf{S} = \sum_{i=1}^{3} w_i \mathbf{S}_i. \tag{9}$$

$\mathbf{S}_i$ represents the normalized similarity matrix calculated using method $i$. The performance of the weightings is evaluated on the training data set aside for validation. The subset of the training data used to compute Eq. 7 is optimised to produce the best retrieval results. Computing these feature weightings only accounts for $\approx 7$ s of the preprocessing time given in Table 2, with the rest of the time spent computing the individual features from the training data to be input into the weight optimization procedure.

Once the best weightings for the training set are obtained, these weightings are then used to combine the similarity matrices computed for the test set, also using Eq. 9.

Results of using HKS, WKS and SA features alone are also given, to show the improvement obtained by this weighted combination.

### 4.6 High-Level Feature Learning for 3D Shapes

The high-level feature learning method for 3D shapes in (Bu et al. 2014a, b) uses three stages (see Fig. 6):

1. Low-level feature extraction: three representative intrinsic features, the *scale-invariant heat kernel signature* (SI-HKS) (Bronstein and Kokkinos 2010), the *shape diameter function* (SDF) (Gal et al. 2007), and the *averaged geodesic distance* (AGD) (Hilaga et al. 2001), are used as low-level descriptors.

2. Mid-level feature extraction: to add the spatial distribution information missing from low-level features, a mid-level position-independent *bag-of-features* (BoF) is first extracted from the low-level descriptors. To compensate for the lack of structural relationships, the BoF is extended to a *geodesic-aware bag-of-features* (GA-BoF), which considers geodesic distances between each pair of features on the 3D surface. The GA-BoF describes the frequency of two geometric words appearing within a specified geodesic distance.

3. High-level feature learning: finally, a deep learning approach is used to learn high-level features from the GA-BoF, which can discover intrinsic relationships using the GA-BoF which provide highly discriminative features for 3D shape retrieval. A stack of *restricted Boltzmann machines* (RBMs) are used, and learning is performed layer by layer from bottom to top, giving a *deep belief network* (DBN) (Hinton et al. 2006). The bottom layer RBM is trained with the unlabelled GA-BoFs, and the activation probabilities of hidden units are treated as the input data for training the next layer, and so on. After obtaining the optimal parameters, the input GA-BoFs are processed layer-by-layer, and the final layer provides the high-level shape features.

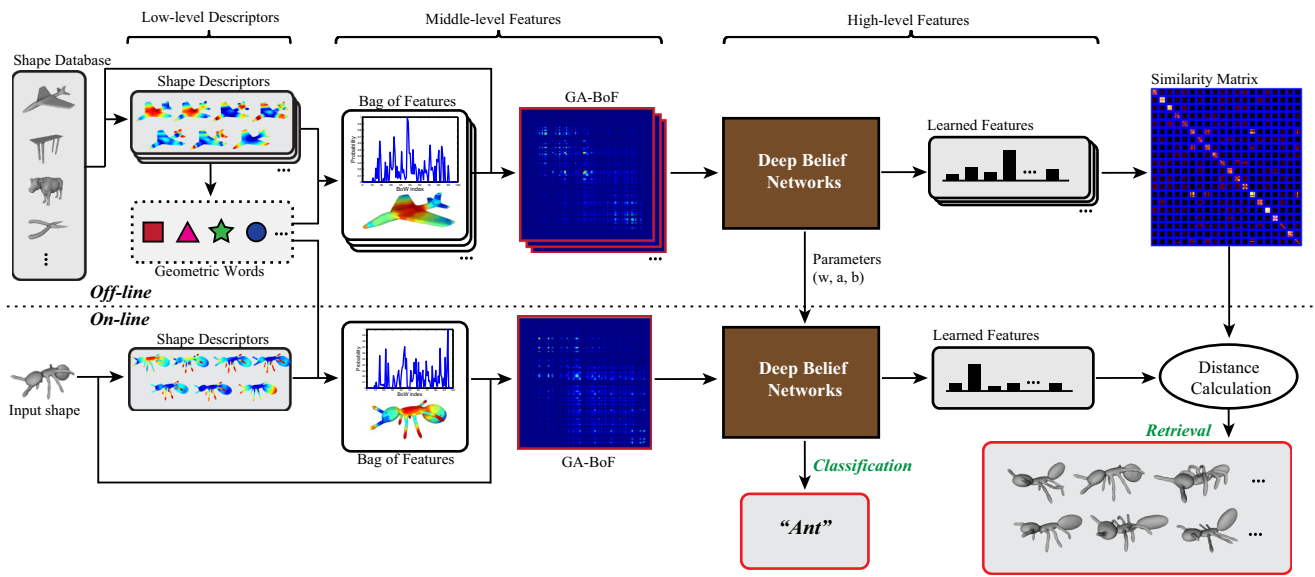

**Fig. 6** Overview of the high-level feature learning method

### 4.7 Bag-of-Features approach with Augmented Point Feature Histograms

*Point feature histograms* (PFH) provide a well-known local feature vector for 3D point clouds, based on a histogram of geometric features extracted from neighbouring oriented points (Rusu et al. 2008). *Augmented point feature histograms* (APFH) improve their discriminative power by adding the mean and covariance of the geometric features.

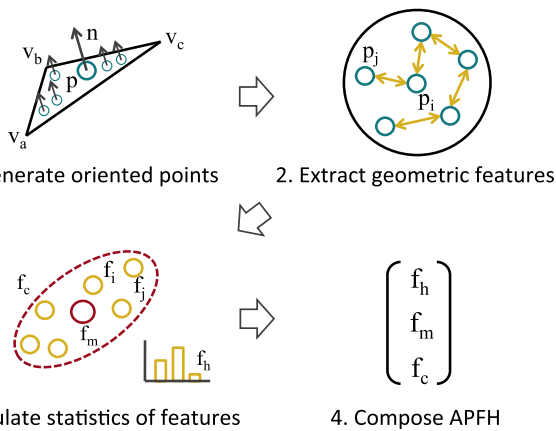

**Fig. 7** Overview of augmented point feature histograms

Because APFH, like PFH, are based on local features, they are invariant to global deformation and articulation of a 3D model.

The APFH approach is illustrated in Fig. 7. The first step is to randomly generate oriented points on the mesh, using Osada's method (Osada et al. 2002). The orientation of each point **p** is the normal vector of the surface at that point.

Next a PFH is constructed for each oriented point. The 4D geometric feature $\mathbf{f} = [f_1, f_2, f_3, f_4]^T$ proposed in Wahl et al. (2003) is computed for every pair of points $\mathbf{p}_a$ and $\mathbf{p}_b$ in the point's $k$-neighbourhood:

$$f_1 = \arctan(\mathbf{w} \cdot \mathbf{n}_b, \mathbf{u} \cdot \mathbf{n}_a), \tag{10}$$

$$f_2 = \mathbf{v} \cdot \mathbf{n}_b, \tag{11}$$

$$f_3 = \mathbf{u} \cdot \frac{\mathbf{p}_b - \mathbf{p}_a}{d}, \tag{12}$$

$$f_4 = d, \tag{13}$$

where the normal vectors of $\mathbf{p}_a$ and $\mathbf{p}_b$ are $\mathbf{n}_a$ and $\mathbf{n}_b$, $\mathbf{u} = \mathbf{n}_a$, $\mathbf{v} = (\mathbf{p}_b - \mathbf{p}_a) \times \mathbf{u}/||(\mathbf{p}_b - \mathbf{p}_a) \times \mathbf{u}||$, $\mathbf{w} = \mathbf{u} \times \mathbf{v}$, and $d = ||\mathbf{p}_b - \mathbf{p}_a||$. These four-dimensional geometric features are collected in a 16-bin histogram $\mathbf{f}_h$. The index of histogram bin $h$ is defined by the following formula:

$$h = \sum_{i=1}^{4} 2^{i-1} s(t, f_i), \tag{14}$$

where $s(t, f)$ is a threshold function defined as 0 if $f < t$ and 1 otherwise. The threshold value used for $f_1$, $f_2$, and $f_3$ is 0, while the threshold for $f_4$ is the average value of $f_4$ in the $k$-neighbourhood.

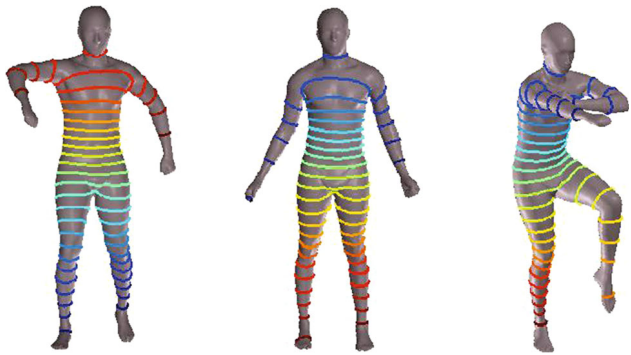

**Fig. 8** Isocontours of the second eigenfunction

The mean $\mathbf{f}_m$ and covariance $\mathbf{f}_c$ of the 4D geometric features is also calculated. The augmented point feature histogram $\mathbf{f}_{APFH}$ comprises $\mathbf{f}_h$, $\mathbf{f}_m$, and $\mathbf{f}_c$. Finally, $\mathbf{f}_{APFH}$ is normalized by power and L2 normalization (Perronnin et al. 2010).

To compare 3D models, the set of APFH features of a 3D model is integrated into a feature vector using the *bag-*

**Table 3** Retrieval results for the *Real* dataset

| Author | Method | NN | 1-T | 2-T | E-M | DCG |
|---|---|---|---|---|---|---|
| Giachetti | APT | **0.830** | 0.572 | 0.761 | 0.396 | *0.826* |
| | APT-trained | *0.910* | *0.673* | **0.848** | **0.414** | *0.874* |
| Lai | HKS | 0.245 | 0.259 | 0.461 | 0.314 | 0.548 |
| | WKS | 0.326 | 0.322 | 0.559 | 0.347 | 0.605 |
| | SA | 0.288 | 0.298 | 0.491 | 0.300 | 0.563 |
| | Multi-feature | 0.510 | 0.470 | 0.691 | 0.382 | 0.708 |
| B. Li | Curvature | 0.083 | 0.076 | 0.138 | 0.099 | 0.347 |
| | Geodesic | 0.070 | 0.078 | 0.158 | 0.113 | 0.355 |
| | Hybrid | 0.063 | 0.091 | 0.171 | 0.120 | 0.363 |
| | MDS-R | 0.035 | 0.066 | 0.129 | 0.090 | 0.330 |
| | MDS-ZFDR | 0.030 | 0.040 | 0.091 | 0.075 | 0.310 |
| C. Li | Spectral Geom. | 0.313 | 0.206 | 0.323 | 0.192 | 0.488 |
| Litman | supDL | *0.775* | **0.663** | *0.859* | *0.421* | **0.857** |
| | UnSup32 | 0.583 | 0.451 | 0.659 | 0.354 | 0.712 |
| | softVQ48 | 0.598 | 0.472 | 0.657 | 0.356 | 0.717 |
| Pickup | Surface area | 0.263 | 0.289 | 0.509 | 0.326 | 0.571 |
| | Compactness | 0.275 | 0.221 | 0.384 | 0.255 | 0.519 |
| | Canonical | 0.010 | 0.012 | 0.040 | 0.043 | 0.279 |
| Bu | 3DDL | 0.225 | 0.193 | 0.374 | 0.262 | 0.504 |
| Tatsuma | BoF-APFH | 0.040 | 0.111 | 0.236 | 0.163 | 0.388 |
| | MR-BoF-APFH | 0.063 | 0.072 | 0.138 | 0.084 | 0.330 |
| Ye | R-BiHDM | 0.275 | 0.201 | 0.334 | 0.217 | 0.492 |
| | R-BiHDM-s | 0.720 | *0.616* | **0.793** | *0.399* | 0.819 |
| Tam | MRG | 0.018 | 0.023 | 0.051 | 0.037 | 0.280 |
| | TPR | 0.015 | 0.024 | 0.057 | 0.050 | 0.288 |

The *1st*, **2nd** and *3rd* highest scores of each column are highlighted

**Table 4** Retrieval results for the *Synthetic* dataset

| Author | Method | NN | 1-T | 2-T | E-M | DCG |
|---|---|---|---|---|---|---|
| Giachetti | APT | **0.970** | 0.710 | 0.951 | 0.655 | 0.935 |
| | APT-trained | *0.967* | **0.805** | **0.982** | **0.692** | *0.958* |
| Lai | HKS | 0.467 | 0.476 | 0.743 | 0.504 | 0.729 |
| | WKS | 0.810 | 0.726 | 0.939 | 0.667 | 0.886 |
| | SA | 0.720 | 0.682 | 0.973 | 0.670 | 0.862 |
| | Multi-feature | 0.867 | 0.714 | *0.981* | 0.682 | 0.906 |
| B. Li | Curvature | 0.620 | 0.485 | 0.710 | 0.488 | 0.774 |
| | Geodesic | 0.540 | 0.362 | 0.529 | 0.363 | 0.674 |
| | Hybrid | 0.430 | 0.509 | 0.751 | 0.520 | 0.768 |
| | MDS-R | 0.267 | 0.284 | 0.470 | 0.314 | 0.594 |
| | MDS-ZFDR | 0.207 | 0.228 | 0.407 | 0.265 | 0.559 |
| C. Li | Spectral Geom. | *0.993* | **0.832** | 0.971 | *0.706* | **0.971** |
| Litman | supDL | 0.963 | *0.871* | 0.974 | **0.704** | *0.974* |
| | UnSup32 | 0.893 | 0.754 | 0.918 | 0.657 | 0.938 |
| | softVQ48 | 0.910 | 0.729 | 0.949 | 0.659 | 0.927 |
| Pickup | Surface area | 0.807 | 0.764 | *0.987* | 0.691 | 0.901 |
| | Compactness | 0.603 | 0.544 | 0.769 | 0.527 | 0.773 |
| | Canonical | 0.113 | 0.182 | 0.333 | 0.217 | 0.507 |
| Bu | 3DDL | 0.923 | 0.760 | 0.911 | 0.641 | 0.921 |
| Tatsuma | BoF-APFH | 0.550 | 0.550 | 0.722 | 0.513 | 0.796 |
| | MR-BoF-APFH | 0.790 | 0.576 | 0.821 | 0.563 | 0.836 |
| Ye | R-BiHDM | 0.737 | 0.496 | 0.673 | 0.467 | 0.778 |
| | R-BiHDM-s | 0.787 | 0.571 | 0.811 | 0.551 | 0.833 |
| Tam | MRG | 0.070 | 0.165 | 0.283 | 0.187 | 0.478 |
| | TPR | 0.107 | 0.188 | 0.333 | 0.216 | 0.506 |

The *1st*, **2nd** and *3rd* highest scores of each column are highlighted

*of-features* (BoF) approach (Bronstein et al. 2011; Sivic and Zisserman 2003). The BoF is projected onto Jensen-Shannon kernel space using the homogeneous kernel map method (Vedaldi and Zisserman 2012). This approach is called BoF-APFH. Similarity between features is calculated using the manifold ranking method with the unnormalized graph Laplacian (Zhou et al. 2011). This approach is called MR-BoF-APFH.

The parameters of the overall algorithm are fixed empirically. For APFH, the number of points is set to 20,000, and the size of the neighbourhood to 55. For the BoF-APFH approach, a codebook of 1200 centroids is generated using $k$-means clustering, and the training dataset is used to train the codebook.

### 4.8 BoF and SI-HKS

This method was presented in Litman et al. (2014). All meshes are down-sampled to 4500 triangles. For each model $\mathcal{S}$ in the data-set, a *scale-invariant heat kernel signature* SI-HKS (Bronstein and Kokkinos 2010) descriptor $\mathbf{x}_i$ is

**Table 5** Retrieval results for the *FAUST* dataset

| Author | Method | NN | 1-T | 2-T | E-M | DCG |
|---|---|---|---|---|---|---|
| Giachetti | APT | **0.960** | **0.865** | **0.962** | **0.700** | **0.966** |
| | APT-trained | *0.990* | *0.891* | *0.984* | *0.711* | *0.979* |
| Lai | HKS | 0.170 | 0.205 | 0.382 | 0.244 | 0.546 |
| | WKS | 0.195 | 0.181 | 0.354 | 0.222 | 0.525 |
| | SA | 0.230 | 0.223 | 0.406 | 0.262 | 0.560 |
| | Multi-feature | 0.350 | 0.226 | 0.379 | 0.246 | 0.573 |
| B. Li | Curvature | 0.805 | 0.644 | 0.777 | 0.558 | 0.853 |
| | Geodesic | – | – | – | – | – |
| | Hybrid | – | – | – | – | – |
| | MDS-R | – | – | – | – | – |
| | MDS-ZFDR | – | – | – | – | – |
| C. Li | Spectral Geom. | 0.555 | 0.255 | 0.369 | 0.252 | 0.611 |
| Litman | supDL | 0.835 | 0.635 | 0.783 | 0.558 | 0.872 |
| | UnSup32 | 0.770 | 0.523 | 0.670 | 0.477 | 0.812 |
| | softVQ48 | 0.730 | 0.426 | 0.551 | 0.387 | 0.748 |
| Pickup | Surface area | 0.545 | 0.509 | 0.818 | 0.544 | 0.763 |
| | Compactness | 0.405 | 0.377 | 0.653 | 0.429 | 0.679 |
| | Canonical | 0.245 | 0.159 | 0.286 | 0.186 | 0.507 |
| Bu | 3DDL | 0.415 | 0.281 | 0.492 | 0.321 | 0.619 |
| Tatsuma | BoF-APFH | 0.890 | 0.652 | 0.785 | 0.559 | 0.886 |
| | MR-BoF-APFH | *0.900* | *0.815* | *0.901* | *0.645* | *0.938* |
| Ye | R-BiHDM | 0.645 | 0.368 | 0.533 | 0.370 | 0.698 |
| | R-BiHDM-s | 0.870 | 0.555 | 0.720 | 0.501 | 0.846 |
| Tam | MRG | – | – | – | – | – |
| | TPR | 0.285 | 0.169 | 0.279 | 0.184 | 0.521 |

The *1st*, **2nd** and *3rd* highest scores of each column are highlighted calculated at every point $i \in \mathcal{S}$. Unsupervised dictionary learning is performed over randomly selected descriptors sampled from all meshes using the SPAMS toolbox (Mairal et al. 2009), using a dictionary size of 32. The resulting 32 atom dictionary $\mathbf{D}$ is, in essence, the *bag-of-features* of this method. Next, at every point, the descriptor $\mathbf{x}_i$ is replaced by a sparse code $\mathbf{z}_i$ by solving the pursuit problem:

$$\min_{\mathbf{z}_i} \frac{1}{2}\|\mathbf{x}_i - \mathbf{D}\mathbf{z}_i\|_2^2 + \lambda\|\mathbf{z}_i\|_1. \qquad (15)$$

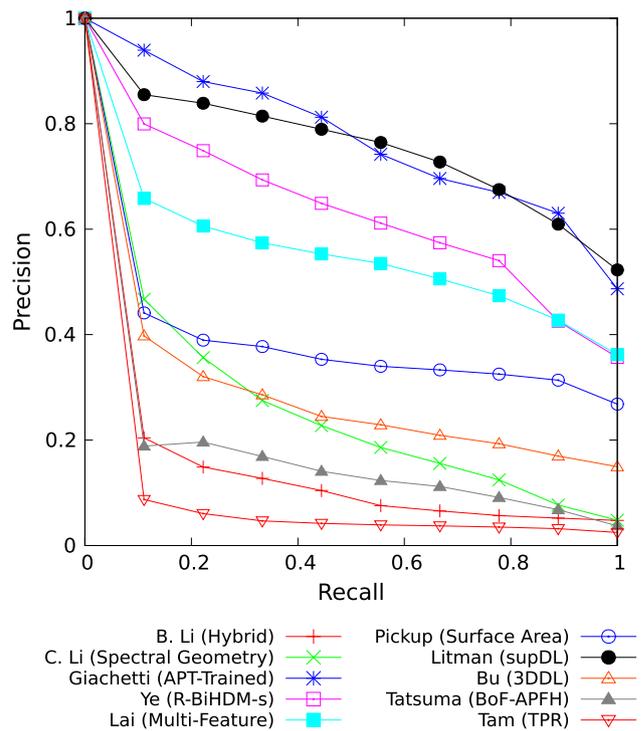

**Fig. 9** Precision and recall curves for the best performing method of each group on the *Real* dataset

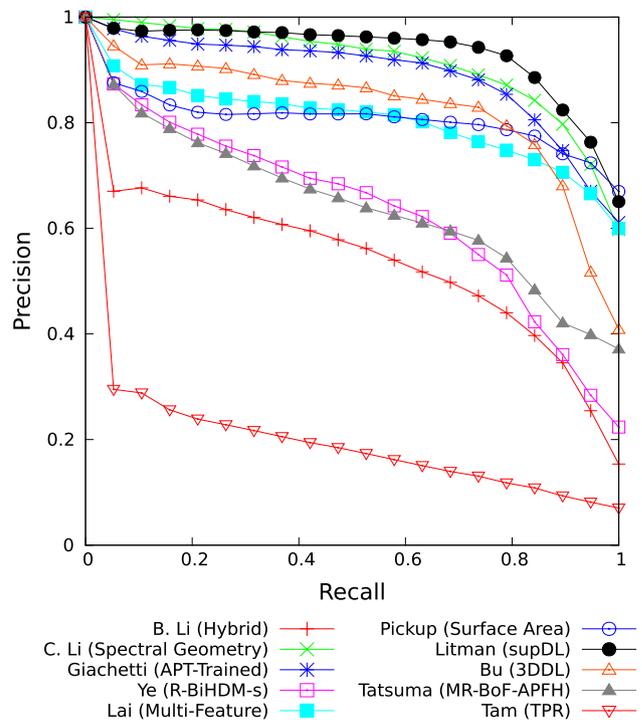

**Fig. 10** Precision and recall curves for the best performing method of each group on the *Synthetic* dataset

The resulting codes $\mathbf{z}_i$ are then pooled into a single histogram using mean pooling $\mathbf{h} = \sum_i \mathbf{z}_i w_i$, with $w_i$ being the area element for point $i$.

The initial $\mathbf{D}$ is determined by supervised training using the training set, using stochastic gradient descent of the loss-function defined in Weinberger and Saul (2009).

The results of three approaches are presented in Sect. 5: the above approach based on supervised training (supDLtrain), and for reference, a method using the initial unsupervised $\mathbf{D}$ (UnSup32). Additionally, the results of a similar unsuper-vised method (softVQ48) used in Bronstein et al. (2011) are also included; it uses $k$-means clustering, with $k = 48$, and soft vector-quantization, instead of dictionary learning and pursuit, respectively.

### 4.9 Spectral Geometry

The spectral geometry based framework is described in Li (2013). It is based on the eigendecomposition of the Laplace-Beltrami operator (LBO), which provides a rich set of eigenbases that are invariant to isometric transformations. Two main stages are involved: (1) *spectral graph wavelet signatures* (Li and Hamza 2013b) are used to extract descriptors, and (2) *intrinsic spatial pyramid matching* (Li and Hamza 2013a) is used for shape comparison.

#### 4.9.1 Spectral Graph Wavelet Signature

The first stage computes a dense spectral descriptor $h(x)$ at each vertex of the mesh $X$. Any of the spectral descriptors with the eigenfunction-squared form reviewed in Li and Hamza (2013c) can be used for isometric invariant representation. Here, the *spectral graph wavelet signature* (SGWS) is used, as it provides a general and flexible interpretation for the analysis and design of spectral descriptors $S_x(t, x) = \sum_{i=1}^{m} g(t, \lambda_i)\varphi_i^2(x)$, where $\lambda_i$ and $\varphi_i$ are the eigenvalues and associated eigenfunctions of the LBO. In the experiments $m = 200$. To capture the global and local geometry, a multi-resolution shape descriptor is obtained by setting $g(t, \lambda_i)$ as a cubic spline wavelet generating kernel. The resolution level is set to 2.

#### 4.9.2 Intrinsic Spatial Pyramid Matching

Given a vocabulary of representative local descriptors $P = \{p_k, k = 1, \ldots, K\}$ learned by $k$-means, the dense descriptor $S = \{s_t, t = 1, \ldots, T\}$ at each point of the mesh is replaced by the Gaussian kernel based soft assignment $Q = \{q_k, k = 1, \ldots, K\}$.

Any function $f$ on $X$ can be written as a linear combination of the eigenfunctions. Using variational characterizations of the eigenvalues in terms of the Rayleigh–Ritz quotient, the second eigenvalue is given by

$$\lambda_2 = \inf_{f \perp \varphi_1} \frac{f'Cf}{f'Af}. \tag{16}$$

The isocontours of the second eigenfunction (Fig. 8) are used to cut the mesh into $R$ patches, giving a shape description which is the concatenation of $R$ sub-histograms of $Q$ with respect to eigenfunction value. To consider the two-sign possibilities in the concatenation, the histogram order is inverted, and the scheme with the minimum cost is considered to be the better match. The second eigenfunction is the smoothest mapping from the manifold to the real line, so this intrinsic partition is stable. Kac (1966) showed that the second eigenfunction corresponds to the sound frequencies we hear the best. Further justification for using the second eigenfunction is given in Li (2013). This approach provably extends the ability of the popular spatial pyramid matching scheme in the image domain to capture spatial information for meshed surfaces, so it is referred to as *intrinsic spatial pyramid matching* (ISPM) Li and Hamza (2013a). The number of partitions is set to 2 here. The dissimilarity between two models is computed as the $L_1$ distance between their ISPM histograms.

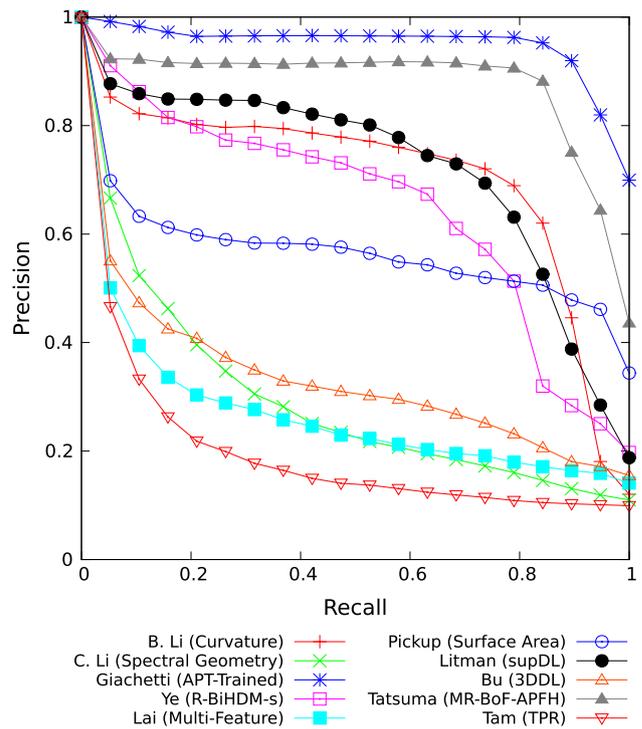

**Fig. 11** Precision and recall curves for the best performing method of each group on the *FAUST* dataset

### 4.10 Topological Matching

This section presents two techniques, topological matching with multi-resolution Reeb graphs, and topological and geometric signatures with topological point rings.

#### 4.10.1 Topological Matching with Multi-resolution Reeb Graphs

The topological matching method was proposed by Hilaga et al. (2001) and is one of the earliest techniques for the retrieval of 3D non-rigid shapes. It begins with the construction of a *multi resolution Reeb graph* (MRG) for each model using integral geodesic distances. Two attributes (local area and length) are calculated for each node of the MRG. The similarity between two MRGs is the sum of the similarity scores between all topologically consistent node pairs. To find these node pairs, the algorithm applies a heuristic graph-matching algorithm in a coarse to fine manner. It first finds the pair of nodes with the highest similarity at the coarsest level, and then finds the pair of child nodes with the highest similarity at the next level. This procedure recurs down both MRGs, and repeats until all possible node pairs are exhausted. It then backtracks to an unmatched highest level node and applies the same procedure again.

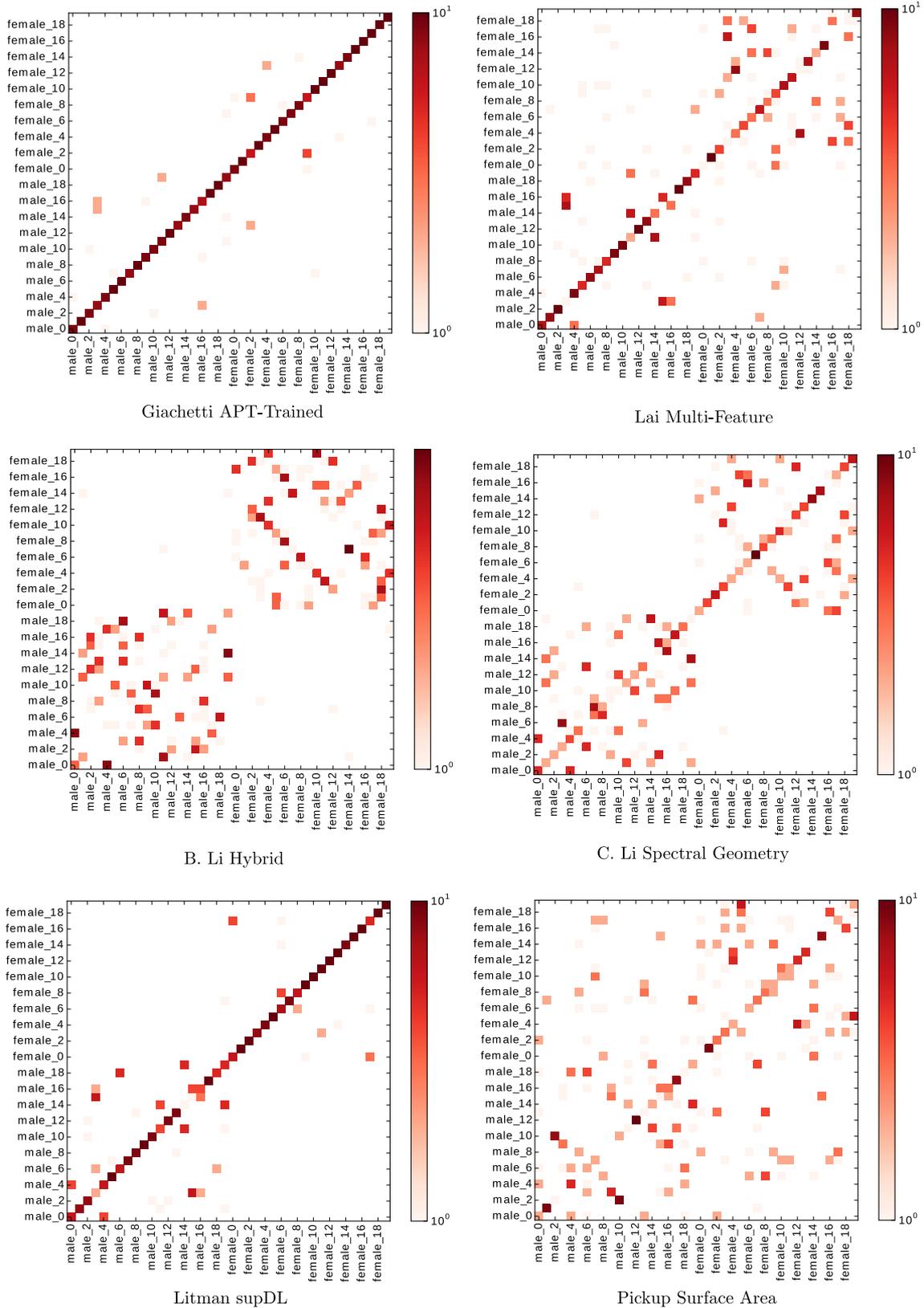

**Fig. 12** Confusion matrix of each method on the *Real* dataset

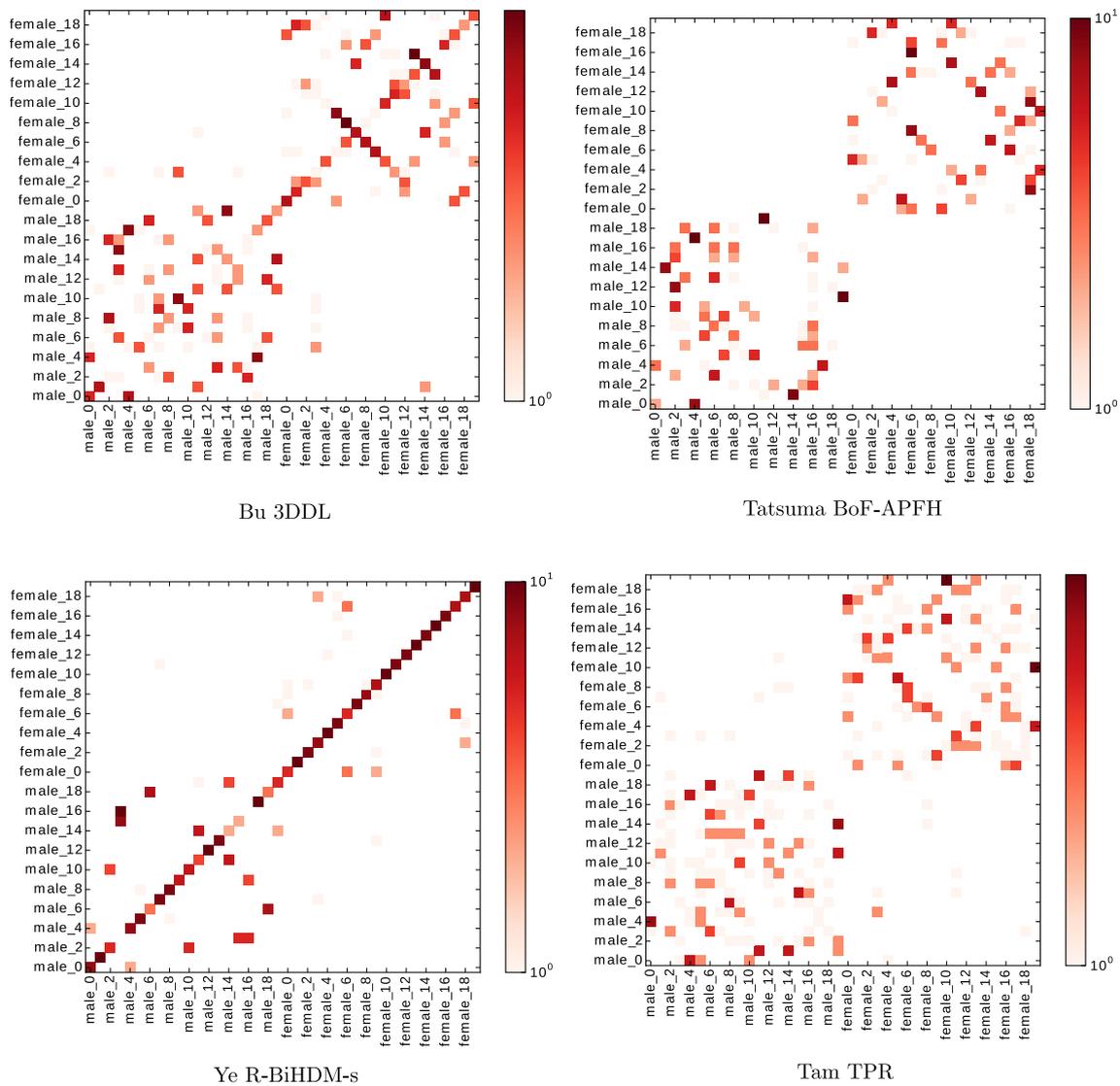

**Fig. 13** Confusion matrix of each method on the *Real* dataset

This method fails on the *FAUST* dataset, as it cannot handle the topological noise present in this data.

### 4.10.2 Topological Point Rings and Geometric Signatures

Topological and geometric signatures were proposed in Tam and Lau (2007). The idea is to define a mesh signature which consists of a set of topologically important points and rings, and their associated geometric features. The earth mover distance (Rubner et al. 2000) is used to define a metric similarity measure between the two signatures of the meshes. This technique is based on skeletal shape invariance, but avoids the high complexity of skeleton-based matching (requiring subgraph-isomorphism). It uses critical points (local maxima and minima of geodesic distance) obtained from a level-set technique to define topological points. With these points, a

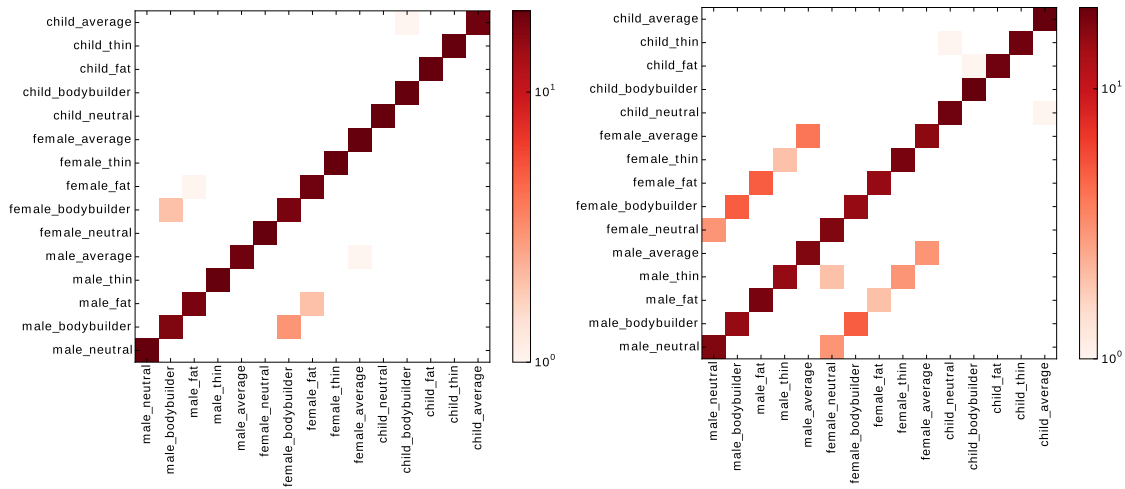

Giachetti APT-Trained                                   Lai Multi-Feature

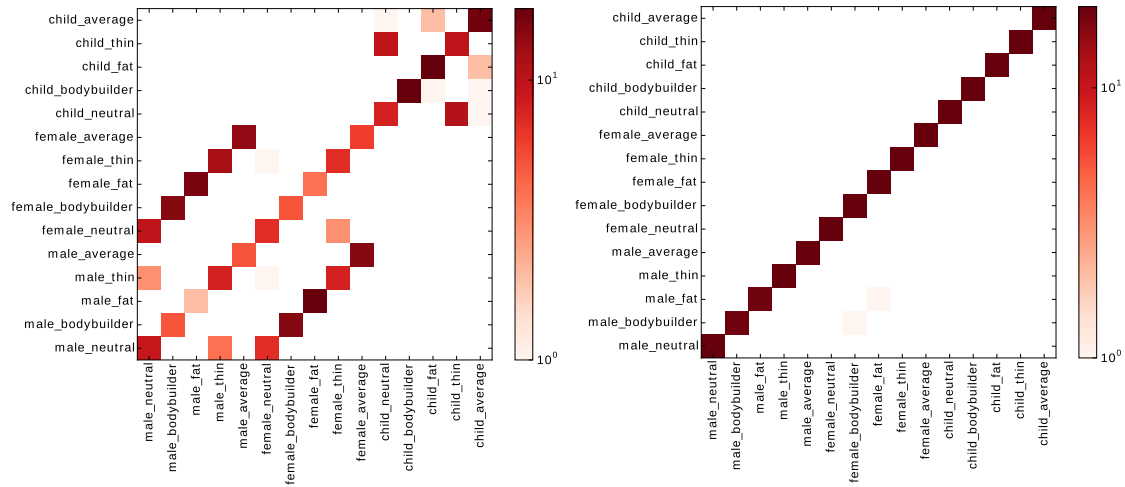

B. Li Hybrid                                            C. Li Spectral Geometry

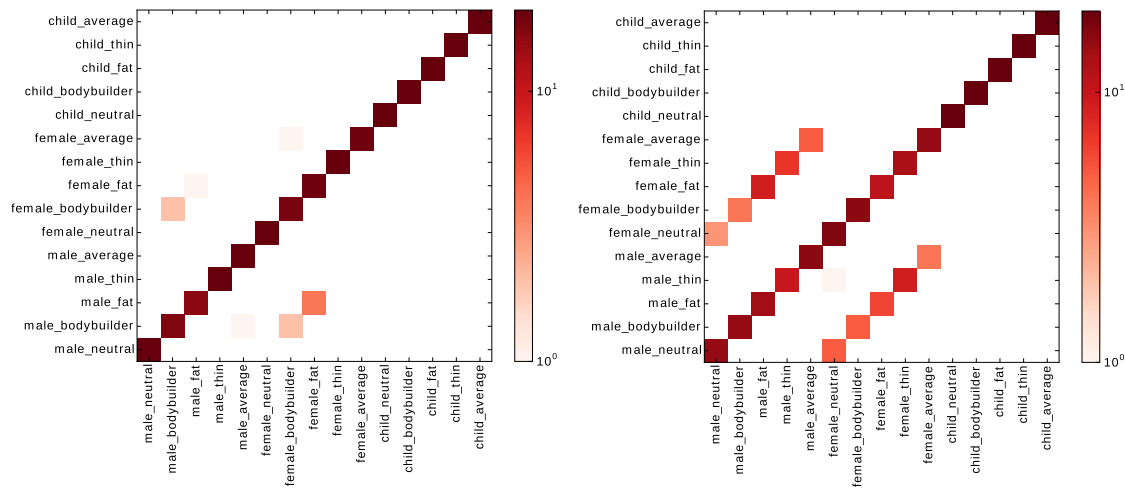

Litman supDL                                           Pickup Surface Area

**Fig. 14** Confusion matrix of each method on the *Synthetic* dataset

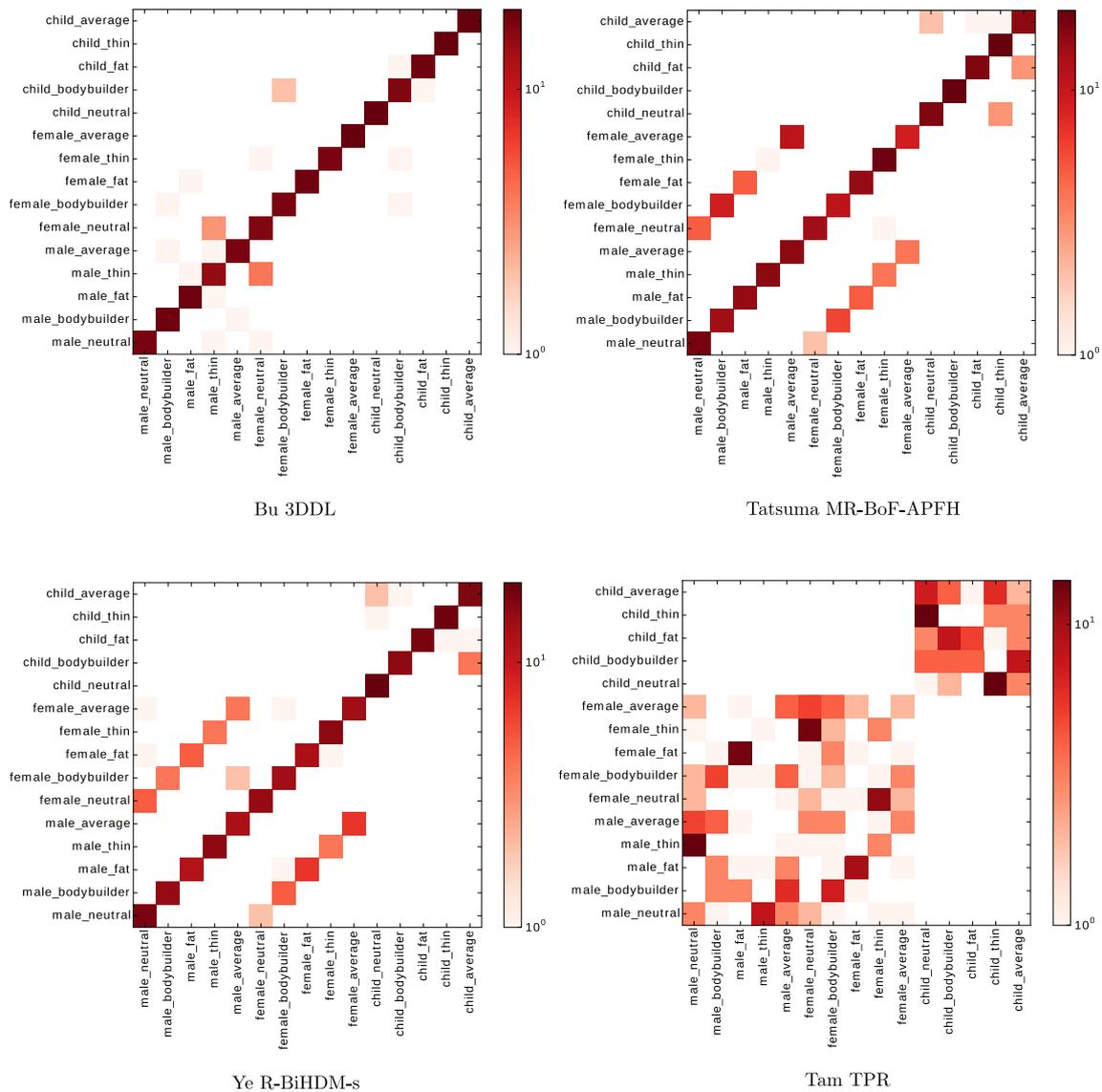

**Fig. 15** Confusion matrix of each method on the *Synthetic* dataset

multi-source Dijkstra algorithm is used to detect geodesic wavefront collisions; the colliding wavefronts give topological rings. For each point or ring, integral geodesic distance and three geometric surface vectors (effective area, thickness, and curvature) are further used to define the final mesh signatures.

## 5 Results

We now present and evaluate the retrieval results for the methods described in Sect. 4, applied to the datasets described in Sect. 2. Retrieval scores are given in Sect. 5.1, then we discuss the results in Sect. 5.2.

### 5.1 Experimental Results

The retrieval task, defined in Sect. 3, was to return a list of all models ordered by decreasing shape similarity to a given query model. Tables 3, 4, and 5 evaluate the retrieval results using the NN, 1-T, 2-T, E-M and DCG measures discussed in Sect. 3. All measures lie in the interval [0, 1], where a higher score indicates better performance.

All methods performed better on the *Synthetic* dataset than the *Real* dataset, with most methods working considerably worse on the *Real* data. Most methods performed somewhere in between these two on the *FAUST* dataset. Figures 9, 10, and 11 show the precision-recall curve for the best performing methods submitted by each participant.

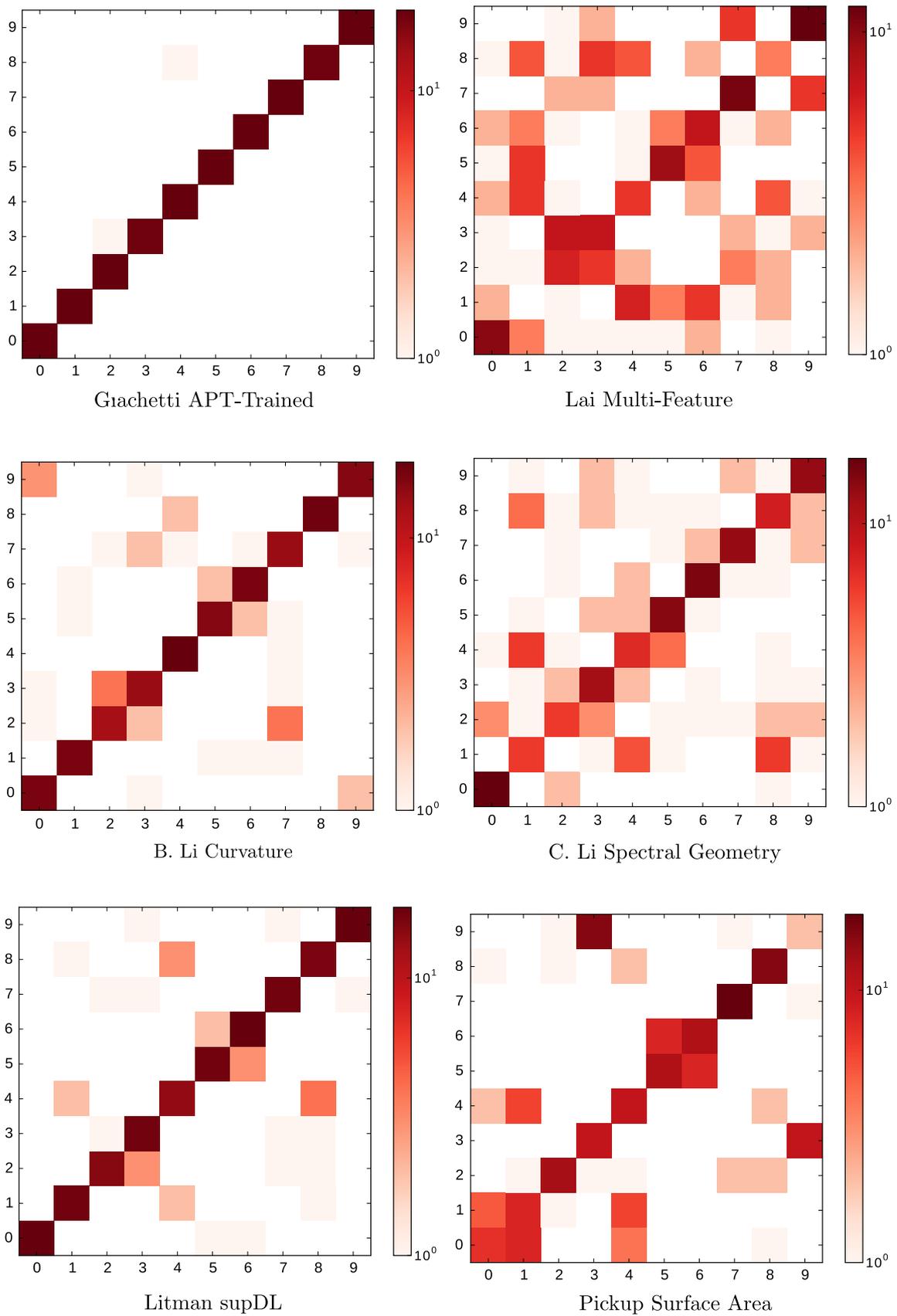

**Fig. 16** Confusion matrix of each method on the *FAUST* dataset

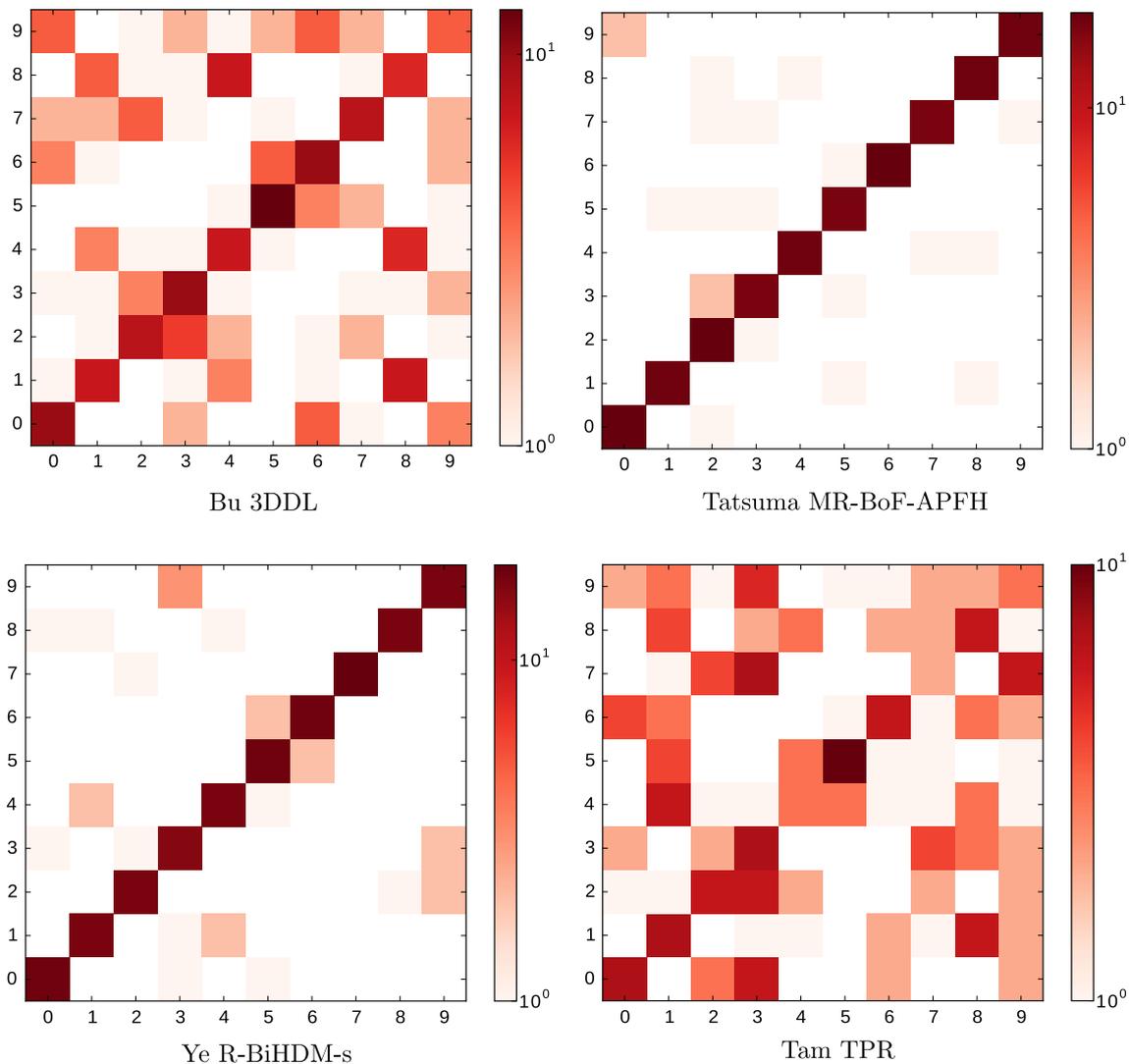

**Fig. 17** Confusion matrix of each method on the *FAUST* dataset

On the most challenging *Real* dataset, supDL by Litman et al., and APT and APT-trained by Giachetti et al. performed best, significantly outperforming other methods, while on the *FAUST* dataset the same is true for the methods by Giachetti et al. and MR-BoF-APFH by Tatsuma and Aono. The performance of different methods is far closer on the *Synthetic* dataset.

We use the precision-recall curves to define which methods perform 'better' than other methods. We say a method performs better than another if its precision-recall curve has higher precision than the other for all recall values. If two curves overlap, we cannot say which method is better.

Figures 12, 13, 14, 15, 16, and 17 show confusion matrices for the best performing methods submitted by each participant for each of the individual classes, for all three datasets.

The corresponding models are rendered in Figures 18, 19, 20, and 21.

### 5.2 Discussion

The results presented in Sect. 5.1 show that performance can vary significantly between different datasets; we may conclude that testing algorithms on one dataset is not a reliable way to predict performance on another.

A possible reason why the different classes in the *Synthetic* data may be more easily distinguished than those in the other datasets is that they were manually designed to be different for this competition, whereas the models in the *Real* and *FAUST* datasets were generated from body scans of human participants taken from an existing dataset, who may or may not have had very different body shapes. There is in fact a much higher similarity between the classes in the *Real* dataset than the other two. This is partly due to the template mesh fitting procedure used in the creation of the *Real* dataset, as it smooths out some of the details present in the scanned meshes. The topological noise present in the *FAUST* dataset also produces an extra challenge.

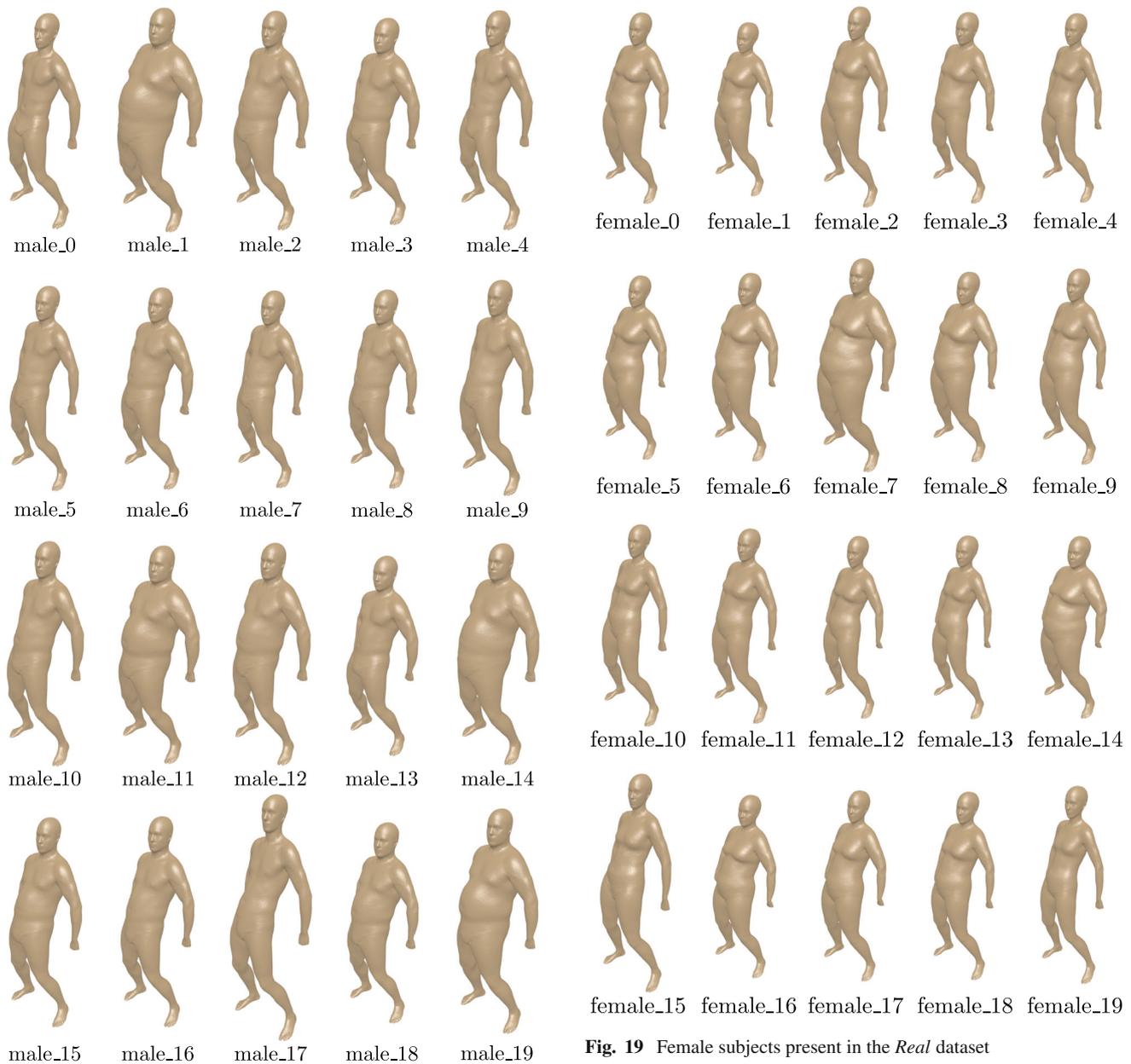

**Fig. 18** Male subjects present in the *Real* dataset

**Fig. 19** Female subjects present in the *Real* dataset

The organisers (Pickup et al.) submitted two very simple methods, surface area and compactness. It is interesting to note that they perform better than many of the more sophisticated methods submitted, including their own. Indeed, surface area is one of the top performing methods on the *Syn-*

*thetic* dataset, with the highest second tier accuracy. These measures are obviously not novel, but they highlight that sophistication does not always lead to better performance, and a simpler and computationally very efficient algorithm may suffice. Algorithms should concentrate on what is truly invariant for each class.

For the *Synthetic* dataset, some methods, including surface area, performed especially well on the child models. This seems to be the same for other methods which are affected by scale. Clearly, methods which take scale into account do not readily confuse children with adults having a similar body shape. The supDL method also exhibits this trend, but claims

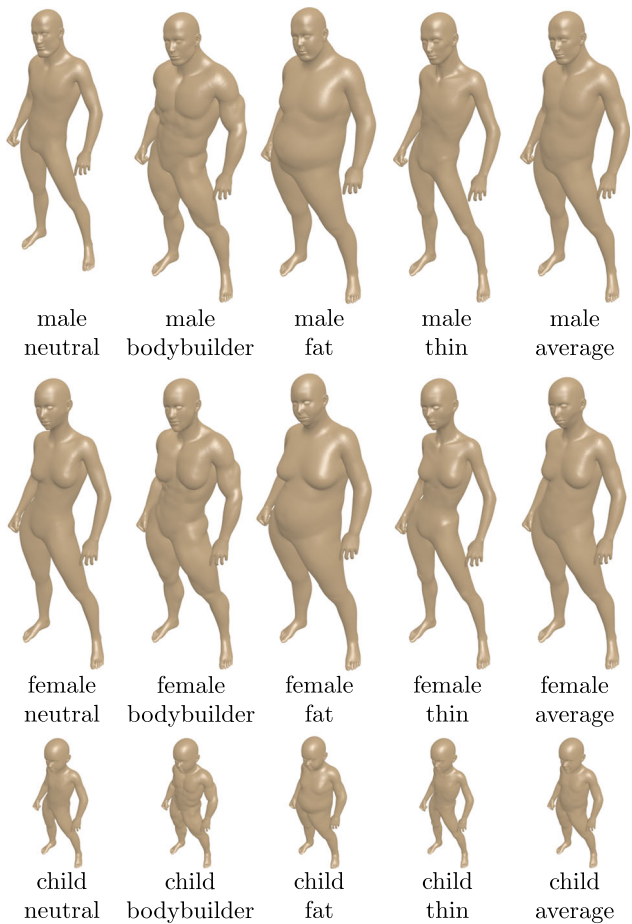

**Fig. 20** Subjects present in the *Synthetic* dataset

male neutral | male bodybuilder | male fat | male thin | male average

female neutral | female bodybuilder | female fat | female thin | female average

child neutral | child bodybuilder | child fat | child thin | child average

**Table 6** The proportion of incorrect nearest neighbour results which are objects with the same pose as the query

| Author | Method | Real | Synthetic | FAUST |
|---|---|---|---|---|
| Giachetti | APT | 0.676 | 0.000 | 0.000 |
| | APT-trained | 0.611 | 0.300 | 0.000 |
| Lai | HKS | 0.109 | 0.025 | 0.060 |
| | WKS | 0.175 | 0.000 | 0.062 |
| | SA | 0.105 | 0.036 | 0.104 |
| | Multi-feature | 0.276 | 0.000 | 0.169 |
| B. Li | Curvature | 0.681 | 0.702 | 0.333 |
| | Geodesic | 0.909 | 0.768 | – |
| | Hybrid | 0.924 | 0.944 | – |
| | MDS-R | 0.969 | 0.927 | – |
| | MDS-ZFDR | 0.905 | 0.861 | – |
| C. Li | Spectral Geom. | 0.807 | 0.000 | 0.371 |
| Litman | supDL | 0.778 | 1.000 | 0.848 |
| | UnSup32 | 0.886 | 0.969 | 0.826 |
| | softVQ48 | 0.758 | 1.000 | 0.685 |
| Pickup | Surface area | 0.112 | 0.017 | 0.154 |
| | Compactness | 0.093 | 0.092 | 0.059 |
| | Canonical | 0.995 | 0.987 | 0.338 |
| Bu | 3DDL | 0.561 | 0.087 | 0.325 |
| Tatsuma | BoF-APFH | 1.000 | 0.993 | 0.909 |
| | MR-BoF-APFH | 0.965 | 0.587 | 0.750 |
| Ye | R-BiHDM | 0.903 | 0.506 | 0.634 |
| | R-BiHDM-s | 0.732 | 0.625 | 0.692 |
| Tam | MRG | 0.947 | 0.953 | – |
| | TPR | 0.967 | 0.892 | 0.594 |

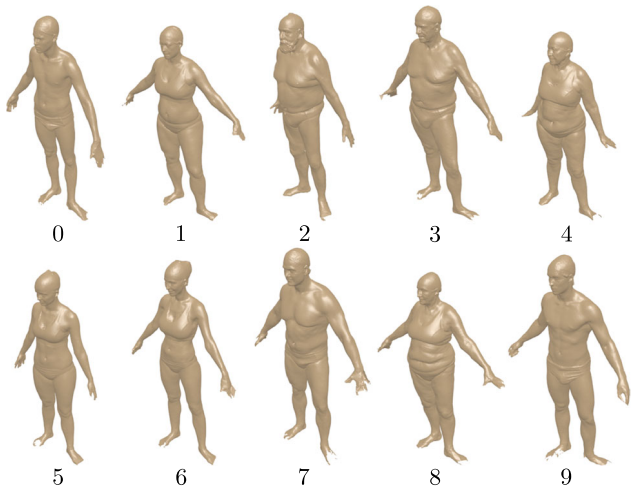

**Fig. 21** Subjects present in the *FAUST* dataset

0 | 1 | 2 | 3 | 4

5 | 6 | 7 | 8 | 9

**Table 7** Correlation coefficient between nearest neighbour retrieval performance, and the percentage of errors which have the same pose as the query

| Real | Synthetic | FAUST |
|---|---|---|
| −0.25 | −0.50 | 0.46 |

to be scale invariant. Ye et al. submitted a scale invariant and a scale dependent version of their algorithm; the corresponding retrieval results demonstrate that a scale dependent method provides significantly improved retrieval accuracy for this dataset.

The APT-trained and supDL methods which performed best on the *Real* dataset, and were amongst the highest performing methods on the *Synthetic* and *FAUST* datasets, both took advantage of the training data. Both participants submitted untrained versions of these methods (APT and UnSup32), which performed worse. This demonstrates the advantage of training.

Table 6 shows the proportion of incorrect nearest neighbour results that share the same pose as the query model. This gives us an idea of how much pose may cause these retrieval errors. In Table 7 we have also presented the correlation

**Table 8** Retrieval results for the *Synthetic* dataset without the child models

| Author | Method | NN | 1-T | 2-T | E-M | DCG |
|---|---|---|---|---|---|---|
| Giachetti | APT | **0.955** | 0.672 | 0.939 | 0.644 | 0.920 |
| | APT-trained | **0.955** | *0.783* | *0.988* | **0.688** | *0.950* |
| Lai | HKS | 0.390 | 0.401 | 0.659 | 0.444 | 0.681 |
| | WKS | 0.730 | 0.626 | 0.912 | 0.635 | 0.838 |
| | SA | 0.610 | 0.591 | 0.961 | 0.644 | 0.816 |
| | Multi-feature | 0.815 | 0.645 | *0.973* | 0.661 | 0.873 |
| B. Li | Curvature | 0.520 | 0.451 | 0.733 | 0.487 | 0.748 |
| | Geodesic | 0.440 | 0.336 | 0.519 | 0.351 | 0.654 |
| | Hybrid | 0.290 | 0.461 | 0.737 | 0.498 | 0.732 |
| | MDS-R | 0.205 | 0.249 | 0.422 | 0.281 | 0.567 |
| | MDS-ZFDR | 0.185 | 0.204 | 0.367 | 0.235 | 0.541 |
| C. Li | Spectral Geom. | *0.990* | **0.808** | 0.962 | *0.698* | *0.963* |
| Litman | supDL | 0.945 | *0.832* | 0.961 | **0.686** | *0.963* |
| | UnSup32 | 0.845 | 0.709 | 0.892 | 0.631 | 0.917 |
| | softVQ48 | 0.870 | 0.657 | 0.926 | 0.630 | 0.900 |
| Pickup | Surface area | 0.710 | 0.651 | **0.981** | 0.664 | 0.853 |
| | Compactness | 0.750 | 0.637 | 0.914 | 0.629 | 0.842 |
| | Canonical | 0.000 | 0.136 | 0.302 | 0.190 | 0.452 |
| Bu | 3DDL | 0.905 | 0.682 | 0.888 | 0.607 | 0.897 |
| Tatsuma | BoF-APFH | 0.405 | 0.517 | 0.726 | 0.510 | 0.768 |
| | MR-BoF-APFH | 0.735 | 0.496 | 0.814 | 0.541 | 0.799 |
| Ye | R-BiHDM | 0.690 | 0.456 | 0.652 | 0.459 | 0.754 |
| | R-BiHDM-s | 0.730 | 0.508 | 0.791 | 0.537 | 0.800 |
| Tam | MRG | 0.060 | 0.151 | 0.270 | 0.176 | 0.474 |
| | TPR | 0.085 | 0.161 | 0.304 | 0.190 | 0.490 |

The *1st*, **2nd** and *3rd* highest scores of each column are highlighted. Most methods show a small drop in performance, compared with the results of the full *Synthetic* dataset

**Table 9** Retrieval results for the *Real* dataset when reduced to ten classes

| Author | Method | NN | 1-T | 2-T | E-M | DCG |
|---|---|---|---|---|---|---|
| Giachetti | APT | **0.945** | 0.813 | **0.951** | *0.437* | *0.943* |
| | APT-trained | *0.968* | *0.870* | 0.974 | 0.438 | *0.961* |
| Lai | HKS | 0.625 | 0.628 | 0.878 | 0.433 | 0.804 |
| | WKS | 0.714 | 0.680 | 0.899 | 0.433 | 0.839 |
| | SA | 0.649 | 0.630 | 0.854 | 0.426 | 0.809 |
| | Multi-feature | 0.825 | 0.775 | 0.948 | *0.437* | 0.900 |
| B. Li | Curvature | 0.281 | 0.232 | 0.391 | 0.253 | 0.528 |
| | Geodesic | 0.273 | 0.265 | 0.442 | 0.277 | 0.553 |
| | Hybrid | 0.299 | 0.279 | 0.458 | 0.287 | 0.565 |
| | MDS-R | 0.207 | 0.215 | 0.353 | 0.236 | 0.510 |
| | MDS-ZFDR | 0.147 | 0.184 | 0.338 | 0.226 | 0.476 |
| C. Li | Spectral Geom. | 0.594 | 0.413 | 0.592 | 0.324 | 0.688 |
| Litman | supDL | **0.931** | *0.878* | *0.980* | *0.439* | **0.958** |
| | UnSup32 | 0.831 | 0.720 | 0.902 | 0.429 | 0.890 |
| | softVQ48 | 0.847 | 0.728 | 0.909 | 0.432 | 0.897 |
| Pickup | Surface area | 0.650 | 0.658 | 0.892 | 0.432 | 0.820 |
| | Compactness | 0.563 | 0.525 | 0.760 | 0.395 | 0.744 |
| | Canonical | 0.006 | 0.041 | 0.191 | 0.161 | 0.367 |
| Bu | 3DDL | 0.582 | 0.540 | 0.794 | 0.414 | 0.759 |
| Tatsuma | BoF-APFH | 0.247 | 0.358 | 0.575 | 0.326 | 0.608 |
| | MR-BoF-APFH | 0.182 | 0.205 | 0.335 | 0.224 | 0.500 |
| Ye | R-BiHDM | 0.614 | 0.458 | 0.682 | 0.377 | 0.730 |
| | R-BiHDM-s | 0.910 | **0.838** | 0.950 | 0.434 | 0.941 |
| Tam | MRG | 0.103 | 0.097 | 0.208 | 0.159 | 0.408 |
| | TPR | 0.100 | 0.129 | 0.265 | 0.197 | 0.431 |

The *1st*, **2nd** and *3rd* highest scores of each column are highlighted coefficient between the nearest neighbour retrieval performance and the percentage of errors having the same pose as the query. We may expect best performing methods to be the most pose-invariant, and therefore produce a strong negative correlation. We find a weak negative correlation for the *Real* dataset, a slightly stronger negative correlation for the *Synthetic* dataset, but a positive correlation for the *FAUST* dataset. Overall this shows that the performance of the method is not a reliable indicator of the pose-invariance of a method. The poses for the *Real* and *Synthetic* datasets are synthetically generated, and therefore are identical. The poses for the *FAUST* dataset are produced from scans of each real human subject imitating each of the poses, and therefore will not be perfectly equal. This may contribute to the very different correlation coefficient for the *FAUST* dataset, shown in Table 7.

Many methods performed significantly better at retrieval on the *Synthetic* dataset. The spectral geometry method of Li et al., which performed poorly on the *Real* and *FAUST* datasets, was one of the best performing methods on the

*Synthetic* dataset. Figures 9 and 10 show that this method fell below the performance of four of the methods analysed using precision and recall on the *Real* dataset and five on the *FAUST* dataset, but was not outperformed by any method on the *Synthetic* dataset. This suggests that there may be features present in the synthetic models which this method relies on to achieve its high performance, yet which are absent in the models within the other datasets. None of the nearest neighbour errors for this method on the *Synthetic* dataset were caused by pose, and therefore this method may be able to extract more pose-invariant features from the *Synthetic* dataset than the other two, which may contribute to its increased performance.

The R-BiHDM-s method submitted by Ye performed better than most methods on the *Real* dataset, but exhibited the smallest performance improvement on the *Synthetic* dataset, and was therefore overtaken by many methods. This may imply that this method performs well at distinguishing global features, but does not take advantage of the extra local detail that is present within the *Synthetic* dataset.

                                                           

The MR-BoF-APFH method by Tatsuma and Aono was a low performer on the *Real* and *Synthetic* datasets, but achieved the second best performance on the *FAUST* dataset. The large increase in performance may be due to the large increase in mesh resolution for this dataset. This was also the only method which did not use the watertight version of the *FAUST* dataset. As this method uses very local features, it may be more robust to the topological noise present in the *FAUST* dataset than other methods.

Figures 12, 13, 14, 15, 16, and 17 show the combined confusion matrices for the three methods with the highest NN score for each dataset. These show that for the *Real* dataset, the methods mostly confuse subjects with other subjects of the same gender. This implies that the difference in body shape due to gender is larger than the difference within gender physiques. The largest confusion on the *FAUST* dataset is also between subjects of the same gender. For the *Synthetic* dataset, these methods exclusively confuse adult subjects with other adults of the opposite gender, but with the same physique (thin, fat, etc.). The child subjects are sometimes confused with other child subjects, but not with adults, presumably due to their smaller size.

Some of the differences in the results between datasets may be caused by the different number of models and classes in each dataset. The *Synthetic* dataset is the only dataset containing models of children. As we have already mentioned, Figures 14 and 15 show that there is less confusion with identifying the child models than the adult models. We therefore show the retrieval results on the *Synthetic* dataset when the child models are ignored (Table 8). These results show that most methods drop slightly in performance, but the overall trends remain the same. The *Real* dataset differs from the other two in that it has a much larger number of classes (40, instead of 15 and 10 for the *Synthetic* and *FAUST* datasets). We therefore generate 100 different subsets of the *Real* dataset, each subset containing a random selection of 10 classes from the original dataset. We perform retrieval on each of these subsets, and average the results over the 100 experiments. The retrieval results are shown in Table 9. The performance of most methods does significantly increase when there are fewer classes, and this demonstrates that the larger number of classes contributes to the increased difficulty of this dataset.

# 6 Conclusions

This paper has compared non-rigid retrieval results obtained by 25 different methods, submitted by ten research groups, on benchmark datasets containing real and synthetic human body models. These datasets are more challenging than previous non-rigid retrieval benchmarks (Lian et al. 2011, 2015), as evidenced by the lower success rates. Using multiple datasets also allows us to evaluate how each method performs on different types of data. Both datasets obtained by scanning real human participants proved more challenging than the synthetically generated data. There is a lot of room for future research to improve discrimination of 'real' mesh models of closely similar objects. We also note that real datasets are needed for testing purposes, as synthetic datasets do not adequately mimic the same challenge.

All methods submitted were designed for generic non-rigid shape retrieval. Our new dataset has created the potential for new research into methods which specialise in shape retrieval of human body models.

**Acknowledgments** This work was supported by EPSRC Research Grant EP/J02211X/1. Atsushi Tatsuma and Masaki Aono were supported by Kayamori Foundation of Informational Science Advancement and JSPS KAKENHI Grant Numbers 26280038, 15K12027 and 15K15992. Zhouhui Lian was supported by National Natural Science Foundation of China Grant Numbers 61202230 and 61472015.